\documentclass[journal,compsoc,cspaper,final]{IEEEtran}
\usepackage{caption}
\usepackage{graphicx}
\usepackage{amsmath}
\usepackage{amssymb}
\usepackage{booktabs}
\usepackage{bm}
\usepackage{bbding}
\usepackage{multirow}
\usepackage[pagebackref,breaklinks,colorlinks]{hyperref}
\usepackage{ragged2e}
\hypersetup{
    colorlinks=true,
    urlcolor=magenta,
    }

\usepackage{enumitem}
\usepackage{xspace}
\newcommand{\ie}{{\emph{i.e.}},\xspace}

\newcommand{\eg}{{\emph{e.g.}},\xspace}
\newcommand{\etc}{etc.}
\newcommand{\etal}{{\emph{et al.}}}
\usepackage[capitalize]{cleveref}
\crefname{section}{Sec.}{Secs.}
\Crefname{section}{Section}{Sections}
\Crefname{table}{Table}{Tables}
\crefname{table}{Tab.}{Tabs.}

\newcommand{\mname}{{Self-EHDRI}}
\newcommand{\mtask}{{E-BL2SH}}
\newcommand{\mnet}{{E-BL2SH}}
\newcommand{\mdata}{{BL2SHD}}
\ifCLASSOPTIONcompsoc
  \usepackage[nocompress]{cite}
\else
  \usepackage{cite}
\fi
\ifCLASSINFOpdf
\else
\fi
\begin{document}
%
\title{HDR Imaging for Dynamic Scenes with Events}
%
%
%
%

\author{Xiaopeng~Li, Zhaoyuan~Zeng, Cien~Fan, Chen~Zhao, Lei~Deng, Lei~Yu
\IEEEcompsocitemizethanks{\IEEEcompsocthanksitem X.P. Li, Z.Y. Zeng, C.E. Fan, C. Zhao, and L. Yu are with the School of Electronic Information, Wuhan University, Wuhan 430072, China. E-mail: \{xiaopengli2014, zhaoyuan.zeng, fce, zhaoc, ly.wd\}@whu.edu.cn.
\IEEEcompsocthanksitem L. Deng is with the Department of Precision Instrument, Tsinghua University, China. E-mail: leideng@mail.tsinghua.edu.cn.
\IEEEcompsocthanksitem The research was partially supported by the National Natural Science Foundation of China under Grants 62271354, 61871297, 61922065, 41820104006, 61871299, 62276151, and the Natural Science Foundation of Hubei Province, China under Grant 2021CFB467.
\IEEEcompsocthanksitem Corresponding authors: Cien Fan and Lei Yu.
}
}

\IEEEtitleabstractindextext{%
\begin{abstract}
\justifying
High dynamic range imaging (HDRI) for real-world dynamic scenes is challenging because moving objects may lead to hybrid degradation of low dynamic range and motion blur. Existing event-based approaches only focus on a separate task, while cascading HDRI and motion deblurring would lead to sub-optimal solutions, and unavailable ground-truth sharp HDR images aggravate the predicament. To address these challenges, we propose an Event-based HDRI framework within a Self-supervised learning paradigm, \ie\ \emph{\mname}, which generalizes HDRI performance in real-world dynamic scenarios. Specifically, a self-supervised learning strategy is carried out by learning cross-domain conversions from blurry LDR images to sharp LDR images, which enables sharp HDR images to be accessible in the intermediate process even though ground-truth sharp HDR images are missing. Then, we formulate the event-based HDRI and motion deblurring model and conduct a unified network to recover the intermediate sharp HDR results, where both the high dynamic range and high temporal resolution of events are leveraged simultaneously for compensation. We construct large-scale synthetic and real-world datasets to evaluate the effectiveness of our method. Comprehensive experiments demonstrate that the proposed \emph{\mname}\ outperforms state-of-the-art approaches by a large margin. The codes, datasets, and results are available at \url{https://lxp-whu.github.io/Self-EHDRI}.

\end{abstract}

}

\maketitle

\IEEEdisplaynontitleabstractindextext

%
\IEEEpeerreviewmaketitle

\IEEEraisesectionheading{\section{Introduction}\label{sec:introduction}}

\IEEEPARstart{D}{ue} to the limited dynamic range of conventional cameras, modern photography in real-world scenarios often suffers from over- or under-exposures, leading to Low Dynamic Range (LDR) images with intensity saturation~\cite{wang2021deep}. High Dynamic Range Imaging (HDRI) is devoted to restoring informative and visually pleasing textures either from single~\cite{lu2009high,chen2020learning,chen2021hdrunet,wang2022kunet} or multi-exposed~\cite{kalantari2017deep,xu2020fusiondn, xu2020mef} LDR images upon specific assumptions, \eg static or slow-moving objects. However, they often face challenges in real-world scenarios with fast and complex non-uniform motions because the input LDR image always suffers from motion blur, leading to weakened textures, rendering frame-based HDRI methods ineffective \cite{yang2023learning}.



\begin{figure*}[!t]
	\centering
\includegraphics[width=1\textwidth]{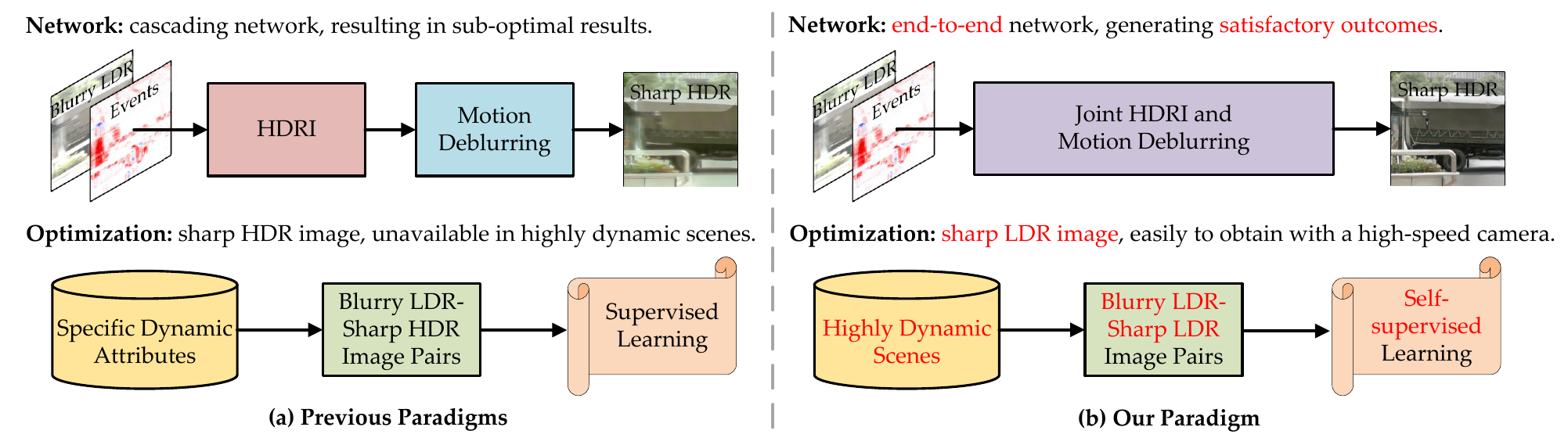} 
	\caption{Previous paradigms vs. our paradigm. (a) Previous paradigms design cascading frameworks in a supervised manner, while ground-truth sharp HDR images can only be obtained in static scenes or dynamic scenes with specific dynamic attributes, \eg actor pose, and controlled lighting conditions (stop-motion capture). When it comes to highly dynamic scenes that contain fast-moving targets, the ground-truth sharp HDR images are usually unavailable, leading to failure in real-world scenarios. (b) We propose an end-to-end and self-supervised framework by devising blurry LDR to sharp LDR conversion, which can handle highly dynamic scenes in the real world without ground-truth sharp HDR images.}
	\label{title}
    \end{figure*}

Recent work has revealed the advantages of event cameras in mitigating the limitations of conventional frame-based HDRI~\cite{han2020neuromorphic, messikommer2022multi} and motion deblurring \cite{purohit2020region, purohit2019bringing}. On the one hand, thanks to the high dynamic range ($>$120 dB) of event cameras, it can compensate for saturated regions in LDR images~\cite{han2020neuromorphic, messikommer2022multi, yang2023learning}. On the other hand, event cameras record brightness changes in microsecond resolution, thus making it feasible to reconstruct latent sharp frames even under nonlinear motions~\cite{haoyu2020learning,xu2021motion,song2022cir}. However, it is still challenging to achieve HDRI for dynamic scenes due to the coupling of {\it hybrid degradation} and the absence of {\it ground truth}, as shown in \cref{title}.

\begin{itemize}
    \item {\bf Coupling of Hybrid Degradation.}
    The event generation model \cite{gallego2020event} that rapidly perceives brightness changes enables the establishment of inter-frame and intra-frame brightness priors, facilitating compensation for HDRI and deblurring tasks, while these priors are degraded when images are both LDR and blurry~\cite{wang2021deep}. In particular, the under- / over-exposed regions in blurry images invalidate the brightness consistency between frames and events, thereby impacting the effectiveness of most deblurring algorithms~\cite{zhang2021event}. Additionally, in scenarios where the LDR image is blurred, overlapping pixel brightness leads to losing reference brightness from regular exposure in event-based HDRI methods, resulting in decreased performance.
    \item {\bf Absence of Ground Truth.}
    Optimization for HDRI in highly dynamic scenes requires sharp HDR images, which are difficult to obtain. Capturing images with alternating exposures for fusion is widely employed to generate ground-truth HDR images while suffering from ghosts in dynamic scenarios~\cite{chen2021hdr,messikommer2022multi}. Although some HDR hardware technologies, such as the Sony IMX490 sensor, demonstrate immunity to ghosting effects, their practical implementation is impeded by the long exposure times spanning, \ie 10 ms~\cite{robidoux2021end, onzon2021neural}. The captured HDR images still contain motion blur when they occur to fast-moving targets. Therefore, it is difficult to generate satisfactory ground-truth images for training in real-world dynamic scenarios, which hinders the efficacy of supervised methods.
\end{itemize}





To achieve HDRI in general dynamic scenes, we propose a Self-supervised Event-based HDRI framework, \ie\ \mname, decoupling the hybrid degradation by learning an Event-based Blurry LDR to Sharp HDR network, \ie\ \mnet\  for joint HDRI and motion deblurring in the LDR domain.
Specifically, the \mnet\ network is comprised of a Dynamic Range Enhancement (DRE) module and a Motion Deblurring (MD) module and is trained end-to-end in a self-supervised manner to avoid sub-optimal results. 
Instead of learning in the HDR domain, we turn to the LDR domain to alleviate the burden stemming from the absence of ground truth sharp HDR images. To this end, a decomposition and composition architecture is proposed to formulate the cross-domain transformations, \ie\ HDR-to-LDR and LDR-to-HDR. The sharp LDR images are then captured with different exposure time intervals, \ie\ EV-2, EV+0, and EV+2, to provide sufficient supervision in the LDR domain, which can be easily obtained with high-speed cameras compared with the sharp HDR references.
A series of cross-domain consistencies are employed for network training, including HDR-LDR and LDR-LDR consistencies for the decomposition network in the LDR domain and LDR-HDR and HDR-HDR consistencies for the composition network and the \mnet\ network in the HDR domain. 

The main contributions of our work are as follows.
\begin{itemize}
\item We propose a self-supervised learning framework to facilitate optimization for sharp HDR image reconstruction in real-world dynamic scenarios without ground-truth sharp HDR images.
\item We propose an efficient event-based neural network to recover sharp HDR images from a blurry LDR frame in an end-to-end manner. To our knowledge, this is the first attempt at leveraging events to address the hybrid degradation of LDR and motion blur.
\item We design a special hybrid imaging system consisting of a conventional camera, an event camera, and a high-speed camera and build a large-scale benchmark dataset composed of real-world blurry LDR images, concurrent event streams, and sharp LDR images.
\end{itemize}

The remainder of this paper is unfolded as follows. \cref{sec:rela} reviews previous works related to our proposed method, including HDRI, motion deblurring, and joint HDRI and deblurring. \cref{problem statement} outlines the problem statement for the Event-based Blurry LDR image to Sharp HDR image (\mtask) task and analyzes its associated challenges. We further describe our proposed self-supervised framework for sharp HDR image reconstruction, \ie\ \mname\ in \cref{Proposed Method} and present implementation of the \mnet\ network in \cref{E-BL2SH network}. Then, a new real-world benchmark dataset is introduced for the \mtask\ task in \cref{dataset}. Finally, we conduct qualitative and quantitative comparisons with several state-of-the-art methods, followed by the ablation studies in \cref{Experiments}. Conclusions are given in \cref{Conclusion}.

\section{Related Work}
\label{sec:rela}
\subsection{HDR Imaging}
Frame-based HDR imaging methods can be roughly classified into single-exposure and multi-exposure approaches according to the number of input exposures. Single-exposure approaches generally design hallucination networks to extract features from unsaturated regions and predict details of over- / under-exposed regions~\cite{banterle2017advanced,eilertsen2017hdr,marnerides2018expandnet,chen2020learning,chen2021hdrunet,wang2022kunet}. However, with the limited information contained in one single image, the reconstructed structures of oversaturated regions are not sufficiently natural under extreme conditions. In contrast, multi-exposure approaches yield better performance with the help of additional information contained in bracketed LDR images at the expense of bringing ghosting effects in dynamic scenarios~\cite{debevec1997recovering,kalantari2017deep,chen2021hdr}. Since ground-truth HDR images are difficult to obtain in real-world scenes, recent works also focus on self-supervised learning for learnable HDRI estimation that alleviates the need for ground-truth HDR labels~\cite{nazarczuk2022self}.
	
Recent methods have revealed the advantages of the event camera in compensating for either single-exposure or multi-exposure HDRI tasks by leveraging its high dynamic range and high temporal resolution. 
Regarding single-exposure HDRI tasks, the high dynamic range of events is excavated to facilitate the restoration of saturated regions in both two-stage~\cite{han2020neuromorphic, han2023hybrid} and end-to-end manners \cite{yang2023learning}. Regarding multi-exposure HDRI tasks, the high temporal resolution of events can be well exploited to alleviate the ghost effects brought by inter-frame motions~\cite{messikommer2022multi}.

However, due to scene requirements for ground-truth image generation, the previous methods generally focus on static or dynamic scenarios with predefined movements, leading to limitations in scenarios with complex motions.


	 
	
\subsection{Motion Deblurring}
In real-world scenarios, LDR images often suffer from motion blur due to dynamic targets or camera ego-motion, which aggravates the ill-posedness of HDRI. 
Conventional frame-based motion deblurring methods attempt to recover the latent sharp frame by estimating the blur kernels or designing specific neural networks, \eg dense neural networks \cite{purohit2020region}, recurrent neural networks~\cite{purohit2019bringing}, and deformable convolutional networks~\cite{zhu2019deformable}. However, they usually assume specific motion patterns, thus leading to performance degradation in real-world scenes with complex motions.
    
Recent works have revealed the advantages of event cameras~\cite{gallego2020event} and leverage the high temporal resolution to compensate for nonlinear motion deblurring. Based on the physical event generation model, some works reconstruct sharp images from a blurry frame with the assistance of its concurrent event streams~\cite{pan2020high}. Since real-world events are noisy and the read-out bandwidth is limited, learning-based methods are further proposed by designing deep learning frameworks to generate a more stable sharp image sequence to avoid degraded results, which can well handle the nonlinear motion in real-world scenes ~\cite{xu2021motion,zhang2022unifying,song2022cir}, even in low-light conditions~\cite{zhou2023deblurring}. Several studies have also utilized events to address the restoration problem in the presence of hybrid degradations, \ie eSL-Net \cite{wang2020event} and eSL-Net++ \cite{yu2023learning} successfully achieve super-resolution from a single motion blurred image that suffers from joint degradation of motion blur and low spatial resolution.
    
However, existing motion deblurring methods only focus on recovering well-exposed blurry images, while images captured by conventional cameras tend to have a low dynamic range \cite{wang2021deep}, which is more challenging due to loss of luminance values in saturated regions.

\subsection{Joint HDR Imaging and Deblurring}
Existing joint HDR imaging and deblurring methods generally focus on blur caused by camera shake. Some studies investigate the problem by formulating a Bayesian framework and adopting the maximum-likelihood approach to reconstruct a sharp HDR image from blurry images with different exposures. Still, it can only handle linear motion~\cite{lu2009high}. To alleviate this, further research exploits the complementarity between sensor exposures and blur, successfully recovering an HDR image from a sequence of multi-exposure LDR images. However, blur caused by fast-moving targets in dynamic scenes is rarely studied. By simultaneously leveraging the high dynamic range and high temporal resolution of events, our method can handle hybrid degradation of LDR and motion blur~\cite{vasu2018joint}. Moreover, our proposed self-supervised learning framework efficiently alleviates the difficulty in obtaining ground-truth sharp HDR images in dynamic scenes and obtains remarkable performance.

\section{Problem Statement}
\label{problem statement}
The task of \mtask\ is to restore the latent sharp HDR irradiance $\mathcal{\hat{I}}_\mathcal{T}\triangleq \{\hat{I}(t)\}_{t\in \mathcal{T}}$ from a single blurry LDR image $L_\mathcal{T}$ captured within the exposure time $\mathcal{T}$ and the concurrent event streams $\mathcal{E}_\mathcal{T} \triangleq \left\{\left(\mathbf{x}_i, t_i, p_i\right)\right\}, {t_i \in \mathcal{T}}$, 
where $\mathbf{x}_i$ represents the pixel location, $t_i$ and $p_i \in\{+1,-1\}$ denote the timestamp and polarity of the triggered event. Thus, we can formulate \mtask\ as follows:
\begin{equation}
\label{eqEBL2SH}
\mathcal{\hat{I}}_\mathcal{T}=\mathcal{G}\left(t, L_\mathcal{T}, \mathcal{E}_{\mathcal{T}}\right), \forall t \in \mathcal{T},
\end{equation}
with $\mathcal{G}$ representing the function of sharp HDR image reconstruction.

\noindent\textbf{Optimization with GT.} Ideally, providing the dataset $\mathcal{D}_{GT}\triangleq \{{L_{\mathcal{T}}}_i, {\mathcal{E}_{\mathcal{T}}}_i, {\mathcal{I}_\mathcal{T}}_i\}_i$ which contains aligned blurry LDR images, event streams, and sharp HDR ground truths, $\mathcal{G}$ can be optimized by minimizing the following loss $\mathcal{L}$ in terms of $\theta$~\cite{kim2019deep, deng2021deep, perez2021ntire, akyuz2020deep}, \ie
\begin{equation}
\label{supervise_op}
    \underset{\theta}{\operatorname{argmin}} \enspace \mathbb{E}_{\mathcal{D}_{GT}}\left\{\mathcal{L}\left(\mathcal{G}_{\theta}(t, L_{\mathcal{T}}, \mathcal{E}_{\mathcal{T}}), \mathcal{I}_\mathcal{T}\right)\right\}.
\end{equation}

However, generating such ground-truth HDR labels for dynamic scenes proves to be time-consuming in practice~\cite{chen2021hdr,messikommer2022multi}, demanding intricate procedures that necessitate manipulation of specific dynamic attributes within the scene (\eg actor pose) and controlled lighting conditions (stop-motion capturing). Regrettably, such controlled environments are typically absent in real-world dynamic scenarios, particularly those involving complex fast-moving targets~\cite{robidoux2021end, onzon2021neural}. Therefore, instead of struggling to create sharp HDR images, we propose migrating optimization from a supervised manner to a self-supervised one, where sharp LDR images are leveraged instead of sharp HDR ones.

\noindent\textbf{Optimization without GT.} Given a GT-free dataset $\mathcal{D}_{SL}\triangleq \{{L_{\mathcal{T}}}_i, {\mathcal{E}_{\mathcal{T}}}_i, {\mathcal{S}_\mathcal{T}}_i\}_i$ with $\mathcal{S}_\mathcal{T}$ representing the sharp LDR observations which are easily acquired by leveraging a high-speed camera, our optimization paradigm aims at learning \mtask\ with these sharp LDR images, \ie
\begin{equation}
\label{self-supervise_op}
    \underset{\theta}{\operatorname{argmin}} \enspace \mathbb{E}_{\mathcal{D}_{SL}}\left\{\mathcal{L}\left(\mathcal{G}_{\theta}(t, L_{\mathcal{T}}, \mathcal{E}_{\mathcal{T}}), \mathcal{S}_\mathcal{T}\right)\right\}.
\end{equation}



Theoretically, assuming that the sharp LDR observations encompass images with various exposures and can overlay the dynamic range of the scene at any timestamp, \mtask\ can first learn the mapping function from a blurry LDR image to a sharp LDR stack with multiple exposures and then generate the sharp HDR image through the fusion of the sharp LDR stack. However, challenges still exist to realize \mtask\ in real-world dynamic scenarios.

\begin{itemize}
\item Sharp LDR observations captured in highly dynamic scenes are difficult to align spatially due to the movement of the targets, which prevents the conduction of well-aligned blurry LDR image and sharp LDR stack, consequently posing challenges for optimizing the network to learn the complete dynamic range and clear content of the scene.
\item  Although the \mtask\ task can be facilitated by sequentially cascading event-based HDRI~\cite{han2020neuromorphic, messikommer2022multi} to event-based motion deblurring~\cite{haoyu2020learning,xu2021motion,song2022cir} approaches. However, sequential learning causes an accumulation of errors~\cite{kim2019multi}, leading to sub-optimal results.
\end{itemize}

\begin{figure*}[!t]
\centering
\includegraphics[width=1\textwidth]{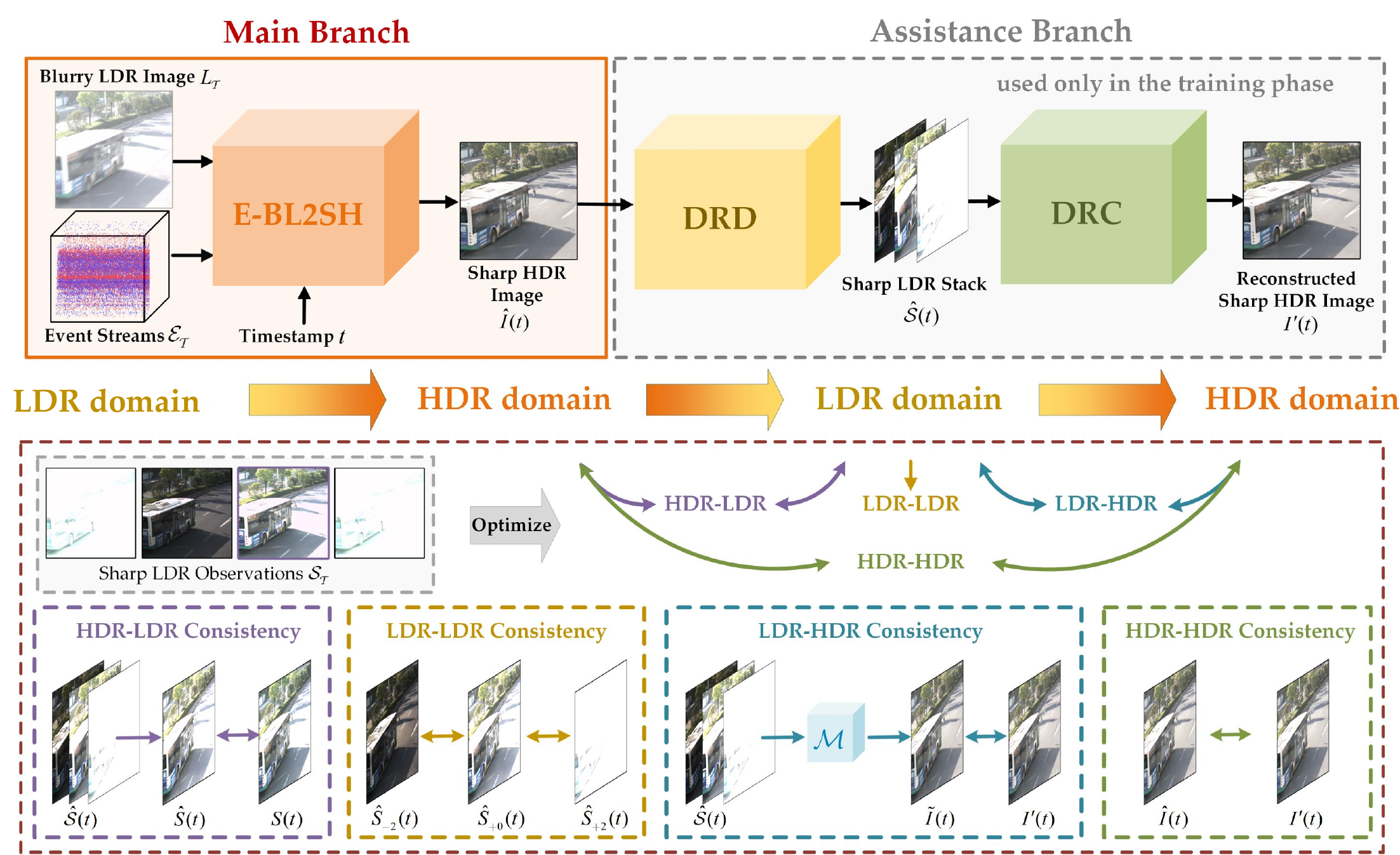} 
\caption{The illustration of our proposed self-supervised learning framework, \ie\ \mname. Top: overall pipeline of \mname. The main branch comprises the \mnet\ network, taking the blurry LDR image and concurrent events as input and generating sharp HDR images at arbitrary timestamps. The assistance branch contains DRD and DRC networks, performing dynamic range decomposition and composition to achieve flexible conversion between HDR and LDR domains. Bottom: the self-supervised consistencies, \ie\ HDR-LDR consistency, LDR-LDR consistency, LDR-HDR consistency, and HDR-HDR consistency, enable the \mtask\ task to be accomplished without ground-truth sharp HDR images.}
\label{fig:main}
\end{figure*}

\section{Learning Strategy}
\label{Proposed Method}
In this section, a novel self-supervised learning framework, \ie\ \mname, is proposed to handle sharp HDR reconstruction in dynamic scenes via events, where the ground-truth sharp HDR images are not required. We present the overall pipeline of \mname\ in \cref{framework}, followed by the self-supervised consistencies in \cref{Self-consistencies}.


\subsection{Overall Pipeline}
\label{framework}
The overall pipeline of the self-supervised learning framework is described in \cref{fig:main}, consisting of two branches, \ie\ {\it the main branch} to reconstruct sharp HDR images from a blurry LDR input and {\it the assistance branch} to facilitate network training without ground-truth sharp HDR images.

The main branch is an Event-based Blurry LDR to Sharp HDR network, \ie\ \mnet, performing joint HDRI and motion deblurring on the blurry LDR input $L_{\mathcal{T}}$ with the aid of event streams $\mathcal{E}_{\mathcal{T}}$ and outputting sharp HDR images $\hat{I}(t)$ of arbitrary timestamps $t$. 
The implementation of the \mnet\ network will be detailed in \cref{E-BL2SH network}. 

The assistance branch can be regarded as a decomposition and composition architecture composed of a Dynamic Range Decomposition (DRD) module and a Dynamic Range Composition (DRC) module. The DRD module transforms the output of \mnet\ from the HDR domain to the LDR domain by decomposing $\hat{I}(t)$ into aligned exposure stack $\hat{\mathcal{S}}(t)$. On the contrary, the Dynamic Range Composition (DRC) module aims at turning LDR images back to the HDR domain by merging the decomposed LDR stack $\hat{\mathcal{S}}(t)$ and restoring the sharp HDR image $I^{\prime}(t)$. Such architecture establishes a cycle consistency with the restored sharp LDR image $\hat{I}(t)$ of the \mnet\ network.

With the combination of the main branch and the assistance branch, our framework can not only recover sharp HDR images but also transform the optimization problem from the HDR domain to the LDR domain, where unaligned sharp LDR observations $\mathcal{S}_\mathcal{T}$ can establish self-supervised consistencies for constraint.

\subsection{Self-supervised Consistencies}
\label{Self-consistencies}
A series of consistencies in both the LDR and HDR domains are devised to ensure that our framework works correctly and interpretably, \ie\ the HDR-LDR and LDR-LDR consistencies to guarantee that the generated LDR stack $\mathcal{\hat{S}}(t)$ contains complete dynamic range and sharp content, and the LDR-HDR and HDR-HDR consistencies to guide our framework in generating high-quality, sharp HDR images in the intermediate process.

\begin{figure}[!t]
\centering
\includegraphics[width=1\columnwidth]{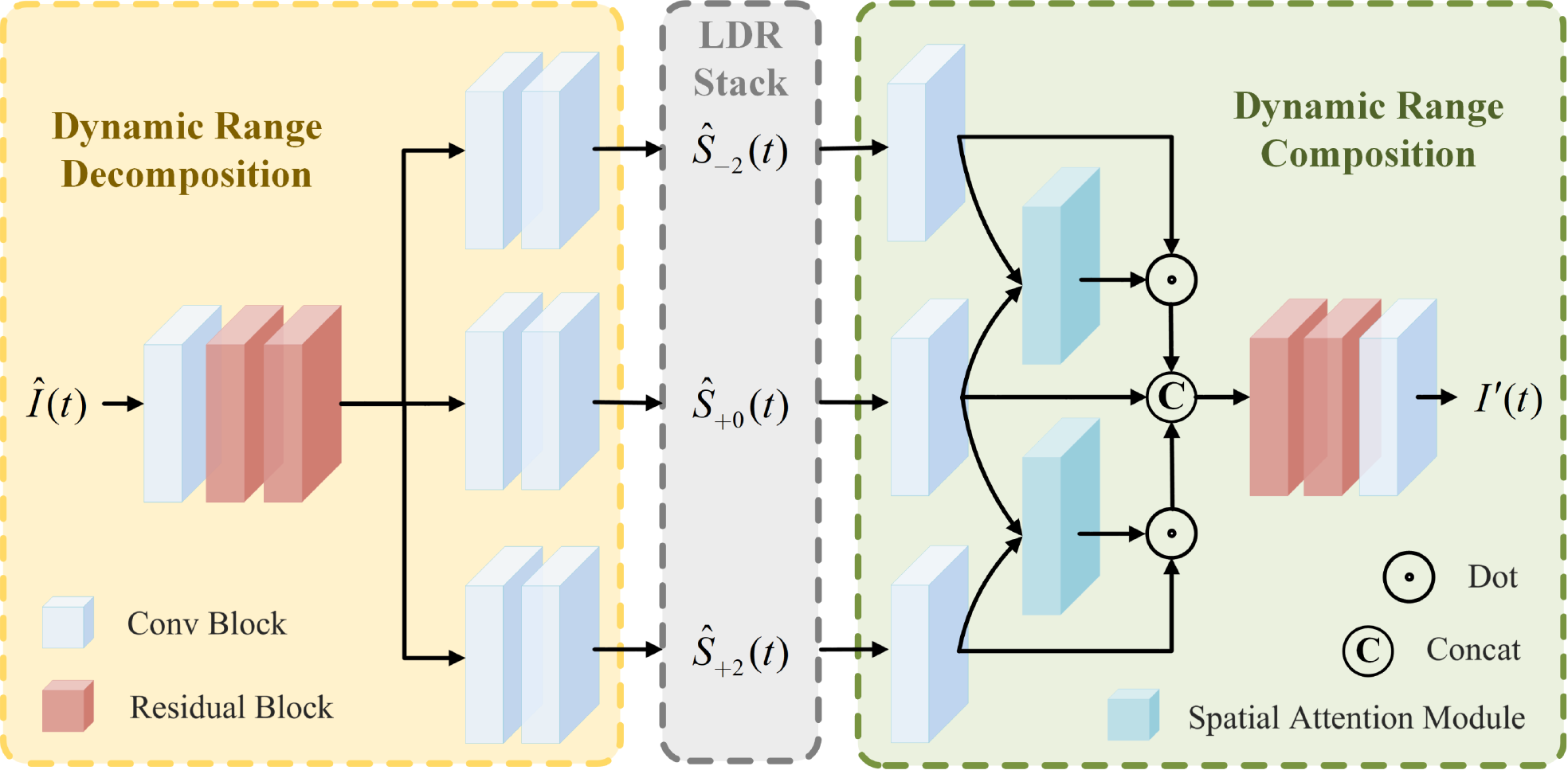} 
\caption{Structure of proposed Dynamic Range Decomposition (DRD) network and Dynamic Range Composition (DRC) network.}
\label{assistance}
\end{figure} 

\noindent\textbf{HDR-LDR Consistency.}
The richness of information within decomposed sharp LDR stack $\hat{\mathcal{S}}_{\mathcal{T}}(t)$ determines the quality of the reconstructed sharp HDR image $I^{\prime}(t)$ and further affects the constraints on $\hat{I}(t)$. 
Therefore, the DRD module is designed to leverage cascaded residual blocks to explore the shallow features of $\hat{I}(t)$ and then transfer its dynamic range to simulate a sharp LDR stack $\mathcal{\hat{S}}_{\mathcal{T}}(t) \triangleq \{\hat{S}_{\text{-2}}(t), \hat{S}_{\text{+0}}(t), \hat{S}_{\text{+2}}(t)\}$ at any given timestamp $t\in \mathcal{T}$,  corresponding to LDR images with different Exposure Values (EV), \ie\ EV-2, EV+0, and EV+2~\cite{chen2021hdr,messikommer2022multi}, as shown in \cref{assistance}.
However, at a given timestamp $t\in \mathcal{T}$, we only capture one single sharp LDR observation, \ie\ $\mathcal{S}_\mathcal{T}(t)\triangleq \{S_{ev}(t)\}, ev\in \{-2,+0,+2\}$, with $S_{ev}(t)$ denoting the sharp LDR image captured with EV=$ev$. 
Thus, one can formulate a loss in the LDR domain,

\begin{equation}
\label{loss HL}
    \mathcal{L}_{HL}  = \sum_{ S_{ev}\in \mathcal{S}_\mathcal{T}} d\left(S_{ev}(t),\mathcal{\hat{S}}_{\mathcal{T}}(t)\right), 
\end{equation}
with $d$ denoting the distance metric between the sharp LDR observation $S_{ev}(t)$ and the decomposed sharp LDR stack $\mathcal{\hat{S}}_{\mathcal{T}}(t)$, \ie\
\begin{equation}
    d\left( S_{ev}(t),\mathcal{\hat{S}}_{\mathcal{T}}(t)\right) \triangleq d_{combo}(S_{ev}(t),\hat{S}_{ev}(t)),
\end{equation}
where $d_{combo}$ represents the combination of L1 loss, perceptual loss \cite{johnson2016perceptual}, and GAN loss \cite{wan2022old}, which is performed at both pixel and feature levels for informative and colorful image generation.

\noindent\textbf{LDR-LDR Consistency.}
A single sharp LDR observation $S_{ev}(t)$ cannot provide supervision for every sharp LDR image in the simulated LDR stack $\mathcal{\hat{S}}_\mathcal{T}(t)$, as described in \cref{loss HL}. Relying on $\mathcal{L}_{HL}$ alone may suffer an imbalance in HDR-to-LDR transformation. 
Thus, we further employ the LDR-LDR consistency to ensure supervision of each LDR observation $S_{ev}(t)$ to the entire simulated sharp LDR stack $\mathcal{\hat{S}}_\mathcal{T}(t)$ by establishing the connection between different LDR exposures. By converting the low-exposure LDR image into the higher one with the brightness conversion function $\mathcal{H}$, the LDR-LDR loss $\mathcal{L}_{LL}$ can be expressed as
\begin{equation}
\begin{aligned}
\mathcal{L}_{LL} & = \|\mathcal{H}(\hat{S}_{-2}(t)) - \hat{S}_{+0} (t)\|_1 \\
& + \|\mathcal{H}(\hat{S}_{+0}(t)) -\hat{S}_{+2}(t)\|_1,
\end{aligned}
\end{equation}
where $\mathcal{H}$ represents linearly multiplying the image brightness by 4 times following \cite{chen2021hdr}.


\noindent\textbf{LDR-HDR Consistency.} 
Although the optimization of $\mathcal{L}_{HL}$ and $\mathcal{L}_{LL}$ facilitates the generation of the sharp LDR stack, there remains an absence of constraints on the \mnet\ network, leading to unpredictable $\hat{I}(t)$. Ideally, we can generate a sharp HDR reference $\tilde{I}(t)$ by performing a conventional multi-exposure image fusion operator $\mathcal{M}$ on $\mathcal{\hat{S}}_\mathcal{T}(t)$ to constrain $\hat{I}(t)$ \cite{chen2021hdr}, \ie
\begin{equation}
\label{EHDRD}
\tilde{I}(t)= \mathcal{M}(\mathcal{\hat{S}}_\mathcal{T}(t)) = \frac{\sum\limits_{ \hat{S}_{ev}\in \mathcal{\hat{S}}_\mathcal{T}} \Phi_{ev}(\hat{S}_{ev}(t)) * \hat{S}_{ev}(t)/n_{ev}}{\sum\limits_{ \hat{S}_{ev}\in \mathcal{\hat{S}}_\mathcal{T}} \Phi_{ev}(\hat{S}_{ev}(t))},
\end{equation}
where $n_{ev} \in \{1,4,16\}$ represents the exposure normalization factor \cite{chen2021hdr} and $\Phi_{ev}$ is employed to calculate the fusion weights for $\mathcal{\hat{S}}_{ev}(t)$, \ie 
\begin{equation}
\label{Phi_i}
\Phi_{ev}(\mathcal{\hat{S}}_{ev}(t))= 1-\max \left(\Lambda_{{ev}}(2*\mathcal{\hat{S}}_{ev}(t)-1), 0\right),
\end{equation}
with $\Lambda_{ev}(z) \triangleq \{z,|z|,-z\}$.

However, due to the instability of $\mathcal{\hat{S}}(t)$ in the early training stages, the merged sharp HDR image $\tilde{I}(t)$ exhibits unreliability, misleading the generation of $\hat{I}(t)$ and further deteriorates $\mathcal{\hat{S}}(t)$. Therefore, we introduce a spatial attention-weighted dynamic range composition (DRC) network, as shown in \cref{assistance}, which enables the fusion process learnable and conditioned by the $\mathcal{L}_{HL}$ and $\mathcal{L}_{LL}$, thereby reducing the impact on the generation of sharp LDR stack. We propose the LDR-HDR consistency to optimize the process by measuring the difference between $\tilde{I}(t)$ and the sharp HDR image $I^{\prime}(t)$ reconstructed through DRC network:
\begin{equation}
\label{efusion}
 \mathcal{L}_{LH}=d_{combo}(\tilde{I}(t), I^{\prime}(t)).
\end{equation}

\begin{figure*}[!t]
\centering
\includegraphics[width=1\textwidth]{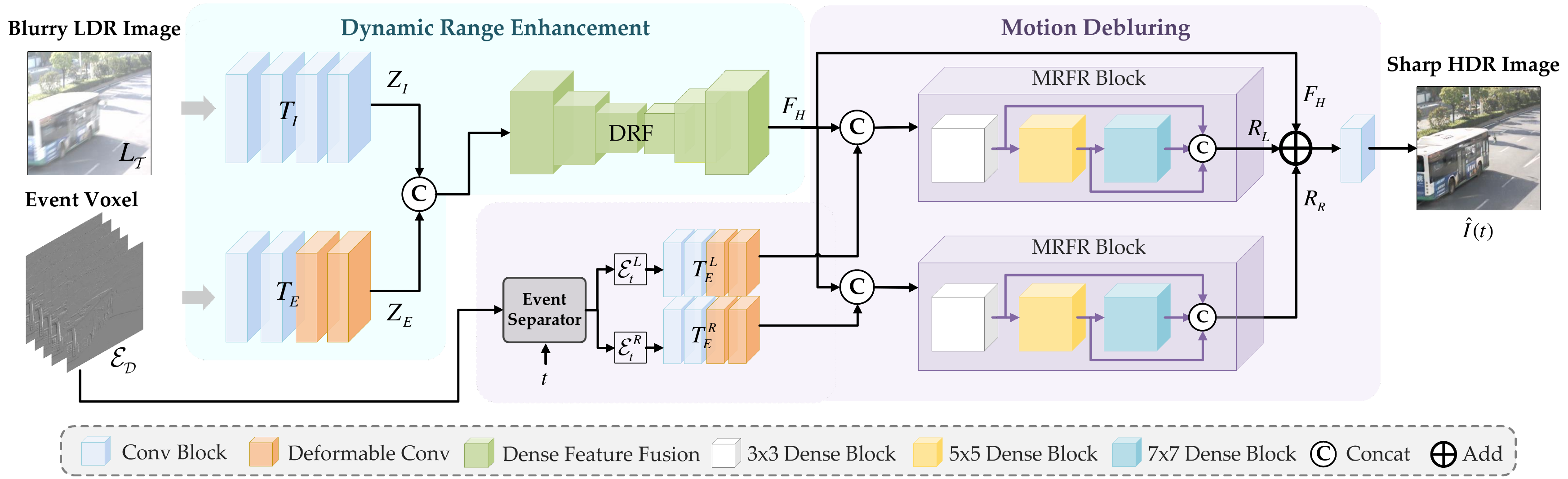} 
\caption{Structure of proposed Event-guided Blurry LDR to Sharp HDR network, \ie\ \mnet, where event streams are embedded into voxel grids with polarity for processability.}
\label{DREDnet}
\end{figure*}

\noindent\textbf{HDR-HDR Consistency.} 
When obtaining the sharp HDR reference $I^{\prime}(t)$, the sharp HDR generation process can be further optimized by minimizing the reconstruction error between $I^{\prime}(t)$ and $\hat{I}(t)$, \ie HDR-HDR loss,
\begin{equation}
\label{rc_loss}
 \mathcal{L}_{HH}=d_{combo}(I^{\prime}(t), \hat{I}(t)).
\end{equation}


Finally, the full loss function for our self-supervised framework is the sum of all sub-loss terms:
\begin{equation}
\mathcal{L}_{total} = \lambda_{1}\mathcal{L}_{HL} + \lambda_{2}\mathcal{L}_{LL} + \lambda_{3}\mathcal{L}_{LH} + \lambda_{4}\mathcal{L}_{HH},
\end{equation}
where $\lambda_{1}$, $\lambda_{2}$, $\lambda_{3}$ and $\lambda_{4}$ are the hyper-parameters that control the trade-off of each term.



\section{Network Architecture}
\label{E-BL2SH network}
In this section, we present our proposed \mnet\ network comprised of a dynamic range enhancement module and a motion deblurring module, as shown in \cref{DREDnet}. To achieve joint HDRI and motion deblurring, we perform \mnet\ in an end-to-end manner following the learning strategy presented in \cref{Proposed Method}. Note that our proposed \mnet\ network is able to generate sharp HDR images at arbitrary timestamps to ensure the feasibility of $\mathcal{L}_{HL}$.


\subsection{Dynamic Range Enhancement Module}
Event cameras and conventional cameras sense brightness in different mechanisms, leading to different data distributions and posing challenges for the fusion process. Therefore, we propose the dynamic range enhancement module to perform multi-modal alignment between the blurry LDR image $L_\mathcal{T}$ and events $\mathcal{E}_\mathcal{T}$ to bridge the distribution gap, then conduct dense fusion to extract high dynamic range features.

For the blurry LDR image $L_\mathcal{T}$, although severely degraded, it retains luminance and color information within unsaturated regions, which can compensate for the details, chrominance, and chroma of the reconstructed images. We conduct an image transform operation $T_I$ to map the degraded LDR image to latent features $Z_I$, where the valuable information can be learned for fusion,
\begin{equation}
    Z_I = T_I(L_\mathcal{T}).
\end{equation}

For event streams $\mathcal{E}_\mathcal{T}$, to make it more processable by convolutional neural networks, we first embed them into a $2m  \times h \times w$ voxel grid $\mathcal{E}_\mathcal{D}$ according to the polarity of events following \cite{zhang2021event}, where $2$ indicates the number of polarity channels and $m$ represents temporal bins. Considering that event representation is inherently sparse, noisy, and lacks details in the low-contrast regions compared with conventional frames, we incorporate deformable convolutions within the event transform module $T_E$. By estimating kernel offsets and modulation masks, $T_E$ attenuates the impact of irrelevant information, \eg event noise, while enhancing the contribution of useful information, \eg gradients. The event feature extraction is defined as,
 \begin{equation}
    Z_E = T_E(\mathcal{E}_\mathcal{D}).
 \end{equation}

The obtained image features $Z_I$ and event features $Z_E$ are integrated by employing the Dynamic Range Fusion module (DRF). We leverage dense feature fusion modules~\cite{dong2020multi} to ensure a sufficient connection between non-adjacent levels of features, resulting in the preservation of spatial information during up/downsampling. Overall, the dynamic range enhancement process can be formulated as,
 \begin{equation}
    F_{H} = \text{DRF}([T_I(L_\mathcal{T}), T_E(\mathcal{E}_\mathcal{D})]),
 \end{equation}
where $F_{H}$ denotes the fused HDR features and $[\cdot, \cdot]$ is the concatenation operation.

\subsection{Motion Deblurring Module}
The goal of the motion deblurring module is to exploit the high temporal resolution of events and generate sharp HDR images from the HDR features $F_{H}$. In order to reconstruct $\hat{I}(t)$ at arbitrary timestamps, the input events $\mathcal{E}_\mathcal{D}$ should be further transformed into a time-specific representation. Specifically, we separate $\mathcal{E}_\mathcal{D}$ into $\mathcal{E}^L_{t}$ within the interval [0, $t$] and $\mathcal{E}^R_{t}$ within the interval [$t$, $\mathcal{T}$], which respectively represent the left event representation and the right event representation of the latent frame at instant time $t$. In this way, sharp HDR images at arbitrary timestamps can be recovered by feeding the corresponding timestamp $t$. Then, two event transform modules $T_E^L$ and $T_E^R$, which share the same weights, are employed to extract event features. 

Following that, two Multiple Receptive Field Residual blocks (MRFR) are introduced to estimate the residual features $R_{L}$ and $R_{R}$ from the HDR features $F_{H}$ and corresponding event features. Specifical to the MRFR block, it comprises three residual dense blocks~\cite{zhang2018residual}, utilizing multi-size filter kernels to extract features with multiple receptive fields. Features generated from each dense block are interconnected through skip connections to keep global and local features well-preserved. The residual estimation can be obtained as follows:
\begin{equation}
    R_{L} = \text{MRFR}([F_{H}, T_E^{L}(\mathcal{E}^{L}_{t})]), 
\end{equation}
\begin{equation}
    R_{R} = \text{MRFR}([F_{H}, T_E^{R}(\mathcal{E}^{R}_{t})]).
\end{equation}

After residual estimation for motion blur, we generate color-informative sharp HDR outcome $\hat{I}(t)$ from $F_{H}$, $R_{L}$, and $R_{R}$ through the addition operation and convolution block:
\begin{equation}
    \hat{I}(t) = conv(F_{H} + R_{L} + R_{R}).
\end{equation}

\begin{figure}[!t]
\centering
\includegraphics[width=1\columnwidth]{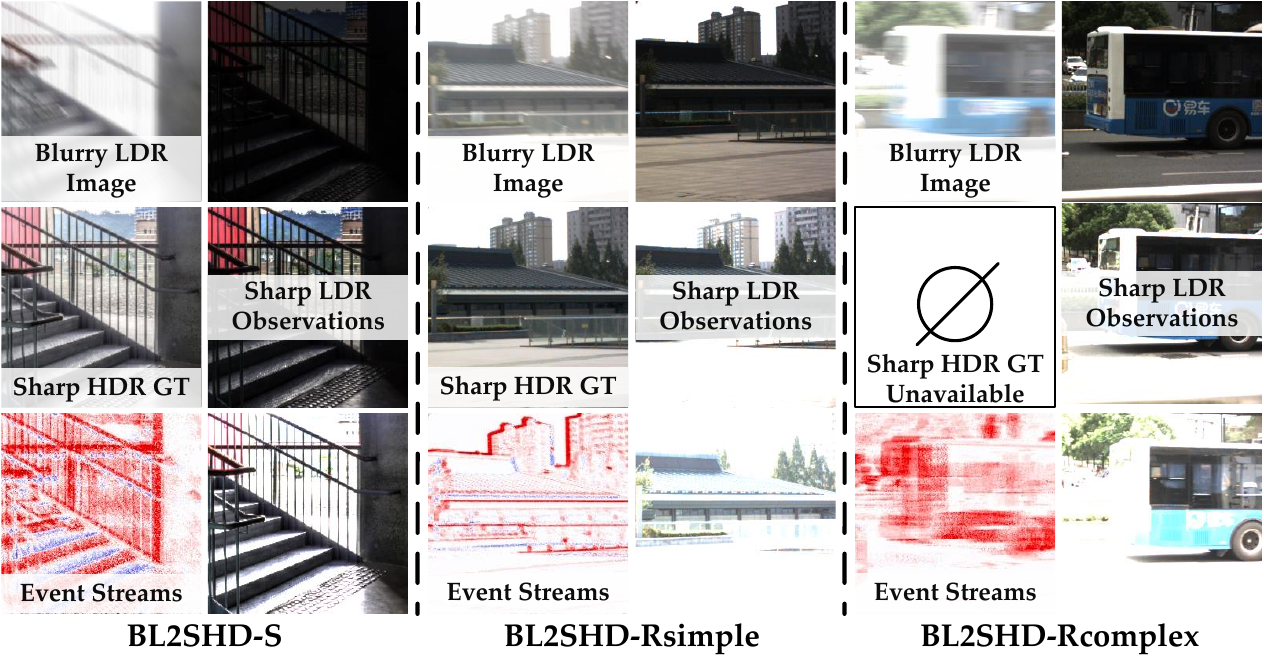} 
\caption{Samples from our proposed \mdata\ dataset, composed of three groups respectively for synthetic, \ie\ \mdata-S, real-world static, \ie\ \mdata-Rsimple, and real-world dynamic scenes \ie\ \mdata-Rcomplex. Note that sharp HDR GT is absent for \mdata-Rcomplex.}
\label{fig:data}
\end{figure}

\section{Datasets}
\label{dataset}
Due to the dearth of available datasets for evaluating \mtask\ methods, we introduce a novel event-based HDRI dataset, \ie\ \mdata, comprised of {\it \mdata-R} captured in real-world scenarios and {\it \mdata-S} synthesized upon existing HDRI dataset, as shown in \cref{fig:data}.

\subsection{Real-world Dataset}
In this subsection, we first introduce our elaborately designed hybrid camera system for capturing data pairs in real-world scenes, followed by the detailed dataset setup. 


\subsubsection{Hybrid Camera System}
\begin{figure}[!t]
\centering
\includegraphics[width=0.95\columnwidth]{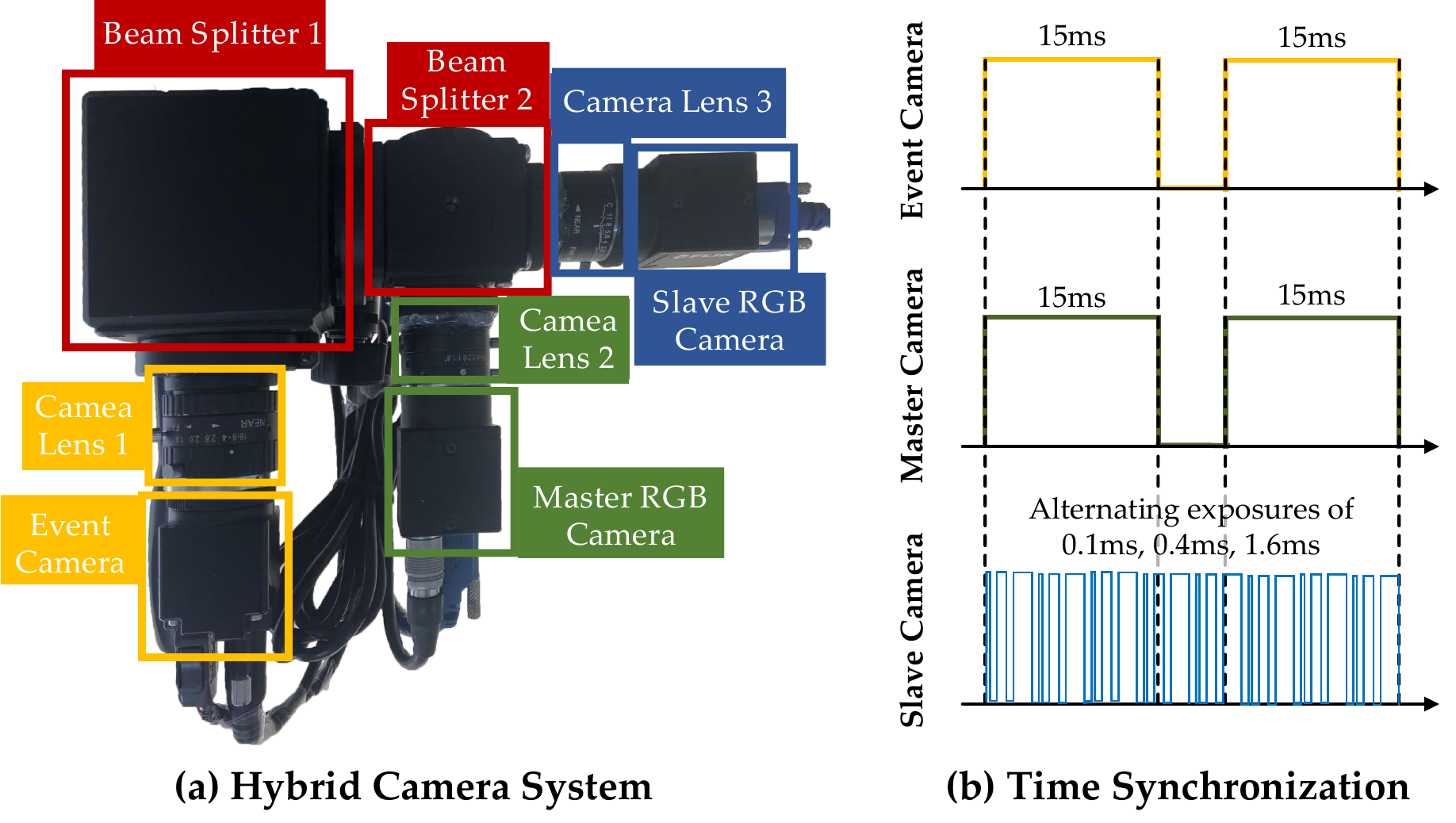} 
\caption{Implementation of our hybrid camera system and its corresponding time synchronization.}
\label{fig:opt}
\end{figure}

As displayed in \cref{fig:opt}, the hybrid camera system is equipped with two beamsplitters connecting three different cameras, including a master camera, \ie\ FLIR BFS-U3-32S4, for capturing real blurry LDR images with the exposure of $15 ms$, an event camera, \ie\ SilkyEvCam, for collecting concurrent event streams, and a high-speed slave camera, \ie\ FLIR BFS-U3-04S2, for capturing sharp LDR frames with alternating exposures of $\{0.1ms, 0.4ms, 1.6ms\}$. Time synchronization is conducted with an electronic system following \cite{zou2021learning}, which generates specially designed triggers to control three cameras to work synchronously, as shown in \cref{fig:opt}. The spatial calibration is also carried out to keep three cameras sharing the same field of view by performing homography estimation \cite{rehder2016extending} between the RGB frame output from the master camera, the RGB frame output from the slave camera, and the intensity images reconstructed from synchronized events~\cite{rebecq2019high}.

\subsubsection{Dataset Setup}
Using this hybrid camera system, we collect a large-scale real-world dataset, \mdata-R, specifically for benchmarking \mtask, under diverse HDR scenarios, including schools, streets, crossroads, and viaducts. By executing rapid and non-linear movements with the hybrid camera system, aligned blurry LDR images, concurrent events, and sharp LDR images are acquired. Furthermore, \mdata\ is divided into {\it \mdata-Rsimple} and {\it \mdata-Rcomplex} according to their motion patterns. 

{\noindent\bf \mdata-Rsimple} is collected in static scenes with camera ego-motion and thus only contains global blurry LDR observations. 
Since all pixels of the captured image are motion-consistent, the ground-truth HDR frames can be generated by global aligning and merging the adjacent multi-exposure sharp LDR images \cite{chen2021hdr}.



{\noindent\bf \mdata-Rcomplex} is collected in dynamic scenes with fast-moving targets, where the captured LDR images encompass global and local non-linear blur, which is more complex to handle. Due to the significant challenges associated with multi-frame registration for fast-moving targets, the ground-truth data for our dynamic scene dataset is currently unavailable.

In this manner, we collect a large-scale dataset containing 9600 aligned data in 48 static HDR scenes and 14400 aligned data in 72 dynamic HDR scenes, as shown in \cref{EBL2SH-RGB}. We divide them into training and testing parts with a ratio of 5:1. Both the blurry LDR image, events, and HDR sequence are resized to $256 \times 256$ for training.

\begin{table}[!t]
    \centering
    \caption{Details of our proposed real-world dataset.} 
    \begin{tabular}{ccccc}
    \hline
\multicolumn{1}{c}{\textbf{Split}} & \multicolumn{1}{c}{\textbf{Scenes}} & \multicolumn{1}{c}{\textbf{\#Seq}} & \multicolumn{1}{c}{\textbf{\#Events (M)}} & \multicolumn{1}{c}{\textbf{\#Pairs}} \\ \hline
\multirow{2}{*}{\textbf{Train}}    & Static  & 40  & 1399.85  & 8000    \\
& Dynamic  & 60 & 1465.54   & 12000  \\ \cline{3-5} 
&  & \textbf{100} & \textbf{2865.39} & \textbf{20000} \\ \hline
\textbf{Test} & Static  & 8  & 425.89   & 1600   \\
& Dynamic & 12  & 322.21  & 2400  \\ \cline{3-5} 
&   & \textbf{20} & \textbf{748.10} & \textbf{4000}  \\ \hline
\textbf{Total}  &  & \textbf{120} & \textbf{3613.49} & \textbf{24000} \\ \hline
    \end{tabular}
    \label{EBL2SH-RGB}
    \end{table}
    
\subsection{Synthetic Dataset}
To evaluate our method more effectively, we construct a synthetic dataset, \ie\ \mdata-S, comprising blurry LDR images, sharp HDR sequences, and their corresponding events, utilizing publicly available HDR datasets, \ie HDR-REAL~\cite{liu2020single}, HDM-HDR-2014~\cite{perez2021ntire}, and DeepHDRVideo~\cite{chen2021hdr}. These datasets contain paired LDR-HDR image pairs, which can be leveraged to synthesize the events through image ego-motion. Specifically, the blurry LDR image is obtained by averaging 13 consecutive sharp clear images, with corresponding HDR images serving as the sharp HDR sequence. We utilize the ESIM simulator \cite{rebecq2018esim} to synthesize concurrent events for each blurry LDR image following the methodology of previous research \cite{wang2020event} and configure the framerate to 150. We decompose the HDR images into LDR images with three alternate exposures following DFHSal \cite{wang2020multi} so that the dataset meets the real scene acquisition conditions. In this manner, we generate 9840 pairs of blurry LDR images, HDR images, and event streams as the simulated static dataset, of which 8856 samples are selected for training, and the remaining 984 samples are left for testing.

\section{Experiments}
 \label{Experiments}
 This section compares our proposed \mname\ approach with existing state-of-the-art methods on the \mdata\ benchmark datasets. First, we introduce the experimental setting, including comparison methods, evaluation strategies, and implementation details in Sec.~\ref{experimental Details}. After that, we provide quantitative and qualitative evaluations on synthetic and real-world data in Sec.~\ref{Evaluation}, respectively, and analyze the results. Besides, the external verification on object detection is performed in Sec.~\ref{Object detection} by taking the \mtask\ methods as a pre-processing step. Finally, comprehensive ablation experiments are conducted in Sec.~\ref{Ablation Study} to verify the effectiveness of individual components of \mname. Detailed datasets, codes, and additional results can be accessed through \url{https://lxp-whu.github.io/Self-EHDRI}.

\subsection{Experimental Settings}
\label{experimental Details}
\subsubsection{Comparison Methods}
Since existing state-of-the-art HDRI methods and deblurring methods cannot deal with blurry LDR images, the proposed \mname\ is compared with cascading approaches, \ie\ {\it HDRI+Deblurring} by implementing HDRI as the pre-operation for deblurring, and {\it Deblurring+HDRI} by implementing deblurring as the pre-operation for HDRI. We adopt frame-based HDRI method KUNet \cite{wang2022kunet} and event-based HDRI method HDRev \cite{yang2023learning} to implement HDRI, and adopt event-based deblurring methods including E-CIR \cite{song2022cir}, EVDI \cite{zhang2022unifying}, and eSL-Net++ \cite{yu2023learning} to achieve deblurring. The task of \mtask\ aims at reconstructing the sharp color HDR sequence, while only EVDI \cite{zhang2022unifying} meets this need. Thus, we report its results with the model provided by the authors trained on color blurry images. As for E-CIR \cite{song2022cir} and eSL-Net++ \cite{yu2023learning}, which work on grayscale images, they are implemented by processing each channel of the RGB image and then combining the outputs to generate color images. 

\begin{table*}[!t]
    \centering
    \caption{Quantitative comparisons to different cascading approaches, \ie\ {\it HDRI-Deblurring} and {\it Deblurring-HDRI} on \mdata-S and \mdata-Rsimple. We highlight the best performance as \textbf{bold}, the second best as \underline{underline}.} 
    \begin{tabular}{cc|ccc|ccc|c}
    \hline
    \multicolumn{2}{c|}{\multirow{2}{*}{Methods}} & \multicolumn{3}{c|}{\mdata-S} & \multicolumn{3}{c|}{\mdata-Rsimple}& \multirow{2}{*}{Runtime$\downarrow$} \\ \cline{3-8}
\multicolumn{2}{c|}{}  & PSNR-$\mu$$\uparrow$ &SSIM-$\mu$$\uparrow$ & HDR-VDP-2$\uparrow$  & PSNR-$\mu$$\uparrow$  & SSIM-$\mu$$\uparrow$ & HDR-VDP-2$\uparrow$  &     \\ \hline
    KUNet & E-CIR & 19.26 & 0.825 & 48.96 & 18.61 & 0.808 & \underline{45.35} & 0.472 \\
    KUNet & EVDI & 19.53 & 0.836 & \underline{50.14} & 18.08 & 0.763 & 44.37 & \textbf{0.023}\\
    KUNet & eSL-Net++ & 19.12 & 0.821 & 48.11 & 18.86 & 0.804 & 44.30 & 0.978 \\
    HDRev & E-CIR & 20.26 & 0.836 & 45.83 & 18.92 & 0.749 & 44.28 & 0.560\\
    HDRev & EVDI & \underline{20.40} & \underline{0.849} & 46.26 & 20.70 & 0.756 & 44.06 & 0.112\\
    HDRev & eSL-Net++ & 20.00 & 0.815 & 43.67 & 20.78 & 0.805 & 42.84 & 1.066\\
    \hline
    E-CIR & KUNet & 19.67 & 0.821 & 48.63 & 17.41 & 0.800 & 44.59 & 0.545\\
    E-CIR & HDRev & 19.60 & 0.840 & 45.37 & \underline{21.37} & \underline{0.809} & 44.47 & 1.693\\
    EVDI & KUNet& 20.10 & 0.830 & 49.86 & 18.23 & 0.762 & 44.20 & 0.096\\
    EVDI & HDRev& 19.74 & 0.843 & 46.17 & 20.02 & 0.728 & 43.78 & 1.244\\
    eSL-Net++ & KUNet & 20.28 & 0.816 & 47.47 & 19.38 & 0.779 & 43.39 & 1.051 \\
    eSL-Net++ & HDRev & 19.47 & 0.830 & 44.97 & 20.48 & 0.772 & 44.02 & 2.199\\
    \hline
    \multicolumn{2}{c|}{Ours} & \textbf{22.34} & \textbf{0.891} & \textbf{52.29} & \textbf{27.80} & \textbf{0.880} & \textbf{49.05} & \underline{0.056} \\\hline
    \end{tabular}
    \label{tab:EBL2SH-S}
    \end{table*}
    
\subsubsection{Evaluation Strategies}
To illustrate the overall HDRI and motion deblurring performance, various metrics are adopted for evaluation from different perspectives, including full-reference and no-reference metrics. 

\noindent\textbf{Full-reference Metrics.} For blurry LDR data in static HDR scenes with motion blur, the evaluation is performed in the $\mu$-law tone-mapping domain, and we utilize the PSNR-$\mu$ \cite{chen2021hdrunet} and SSIM-$\mu$ metrics to quantify the difference between the reconstructed images and the ground-truth sharp HDR sequence. Moreover, to assess the performance in the linear domain, we employ the specific HDRI metric HDR-VDP-2 \cite{mantiuk2011hdr}, which effectively illustrates the visibility and overall quality of HDR images.

\noindent\textbf{No-reference Metrics.} For blurry LDR data in dynamic HDR scenes with motion blur, we adopt the no-reference metrics of Average Gradient (AG) \cite{cui2015detail}, Spatial Frequency (SF) \cite{eskicioglu1995image}, NIQE \cite{mittal2012making}, CLIP-IQA+ \cite{wang2022exploring}, MUSIQ \cite{ke2021musiq}, MANIQA \cite{yang2022maniqa}, and LIQE \cite{zhang2023blind} for evaluation. AG and SF illustrate the gradient information and overall activity in the spatial domain, respectively. NIQE calculates the no-reference image quality score for the input image, which compares the input to a default model generated from natural scene images, and a lower score indicates better perceptual quality. CLIP-IQA+, MUSIQ, MANIQA, and LIQE are trained on conventional no-reference IQA datasets, \ie KonIQ-10k \cite{hosu2020koniq}, PaQ-2-PiQ \cite{ying2020patches} \etc, which are able to measure the sharpness, colorfulness, noisiness, contrast, and quality of the input images. For these metrics, larger values indicate better results.  

\subsubsection{Implementation Details}
\label{Implementation Details}
We implement our model in the Pytorch platform and choose ADAM \cite{kingma2014adam} as the optimizer. We train our proposed \mname\ with NVIDIA GeForce RTX 4090 GPUs for 200 epochs,  set the initial learning rate to 0.0002, and decay the learning rate linearly from 100 epochs to the end. The hyper-parameters \{$\lambda_{1}$, $\lambda_{2}$, $\lambda_{3}$, $\lambda_{4}$\} are set as: \{1, 1, 1, 1\}. Due to the relatively easier implementation of E-BL2SH in static scenes, our training approach involves a sequential process. Initially, our framework is pre-trained on \mdata-Rsimple in a self-supervised manner, prioritizing learning a reliable DRC network since blur caused by camera ego-motion are easier to handle than those caused by dynamic targets. Subsequently, we stabilize the parameters of the DRC network and proceed with the comprehensive training on the complete datasets,  generalizing \mnet\ and DRD networks to dynamic scenarios.

\subsection{Comparison with State-of-the-art Methods}
\label{Evaluation}
\begin{figure*}[!t]
\centering
\includegraphics[width=1\textwidth]{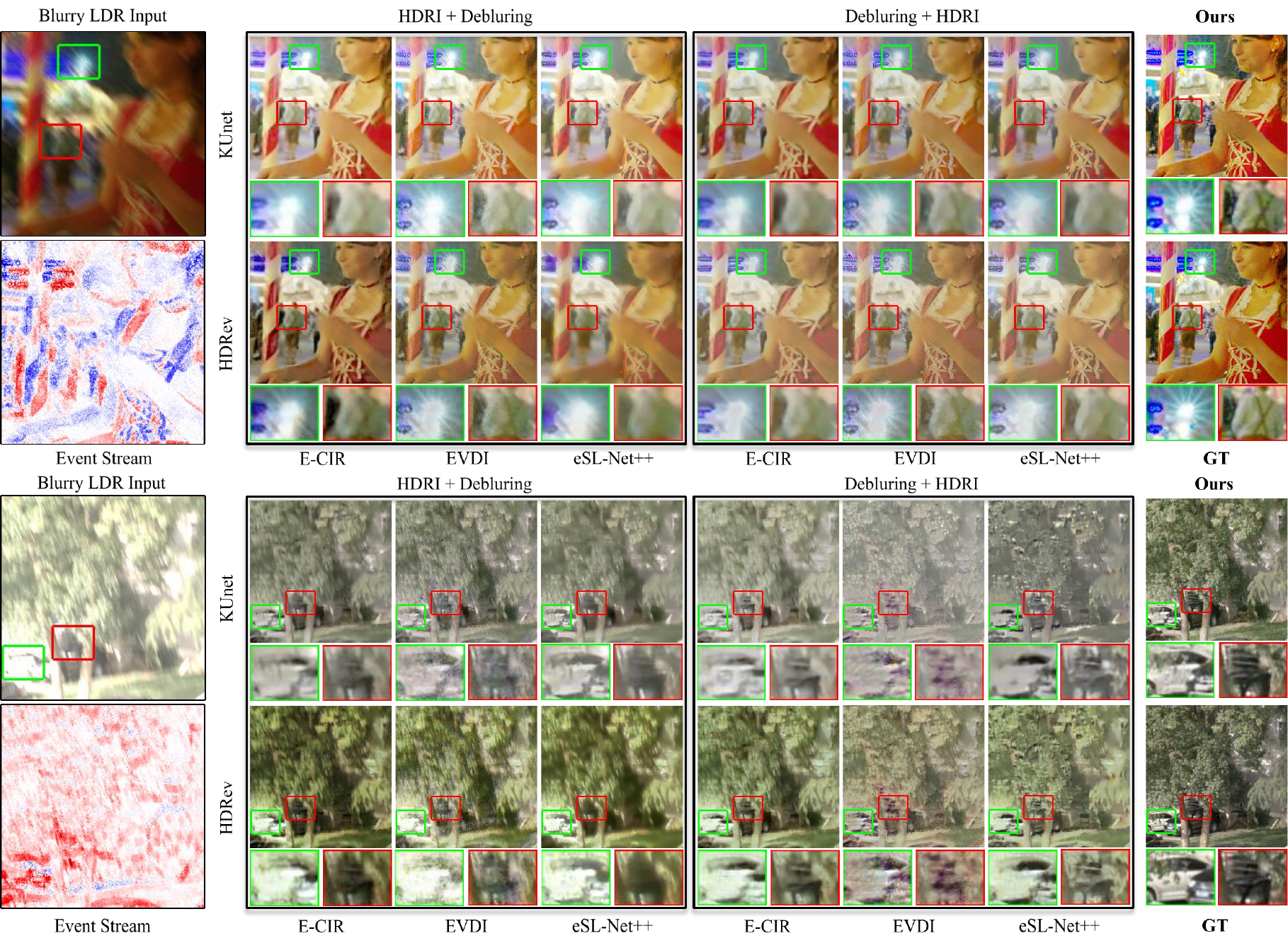} 
\caption{Qualitative results on \mdata-S and \mdata-Rsimple. The combination of state-of-the-art HDRI methods and deblurring methods obtains the compared results. We present several sharp HDR reconstructions that are tone-mapped using the method proposed by Reinhard \etal~\cite{reinhard2002photographic}.}
\label{fig:static}
\end{figure*}

 \begin{table*}[htbp]
    \centering
    \caption{Quantitative comparisons to different cascading approaches, \ie\ {\it HDRI-Deblurring} and {\it Deblurring-HDRI} on real-world dynamic dataset \mdata-Rcomplex. We highlight the best performance as \textbf{bold}, the second best as \underline{underline}.} 
    \begin{tabular}{cc|ccccccc}
    \hline
    \multicolumn{2}{c|}{Methods} & \multicolumn{1}{c}{AG$\uparrow$} & \multicolumn{1}{c}{SF$\uparrow$} & \multicolumn{1}{c}{NIQE$\downarrow$} & \multicolumn{1}{c}{CLIP-IQA+$\uparrow$} & \multicolumn{1}{c}{MANIQA$\uparrow$}& \multicolumn{1}{c}{MUSIQ$\uparrow$} & \multicolumn{1}{c}{LIQE$\uparrow$} \\ \hline
    KUNet & E-CIR &4.833 & 12.336 & 6.347 & \underline{0.407} & \underline{0.505} & 65.051 & 1.747 \\
    KUNet & EVDI &6.364 & 12.637 & 6.046 & 0.390 & 0.449 & 70.617 & 1.679 \\
    KUNet & eSL-Net++ &4.380 & 18.380 & 6.811& 0.375 & 0.428 & 65.416 & 1.588 \\
    HDRev & E-CIR & 6.030 & 15.450 & 5.763 & 0.386 & 0.499 & 65.805 & 1.814 \\
    HDRev & EVDI &\underline{6.926} & 19.688 & 5.189 & 0.377 & 0.462 & 70.920 & 1.742 \\
    HDRev & eSL-Net++ & 4.661 & 13.737 & 7.805 & 0.342 & 0.382 & 62.837 & 1.554 \\
    \hline

    E-CIR & KUNet & 3.754 & 9.751 & 7.299 & 0.395 & 0.471 & 66.095 & 1.717 \\
    E-CIR & HDRev & 5.624 & 14.623  & 5.421 & 0.386 & 0.477 & 67.240 & \underline{1.945} \\
    EVDI & KUNet & 4.295 & 12.655 & 5.452 & 0.372 & 0.423 & 70.368 & 1.523 \\
    EVDI & HDRev & 6.922 & \textbf{21.100} & \underline{4.865} & 0.375 & 0.413 & 71.214 & 1.747 \\
    eSL-Net++ & KUNet & 3.923 & 11.665 & 6.716 & 0.368 & 0.497 & 69.917& 1.715 \\
    eSL-Net++ & HDRev & 5.498 & 16.465 & 5.365 & 0.402 & 0.475 & \underline{71.796} & 1.853 \\
    \hline
    \multicolumn{2}{c|}{Ours} & \textbf{7.193} & \underline{19.830} & \textbf{4.662} & \textbf{0.518} & \textbf{0.605} & \textbf{71.899} & \textbf{3.134} \\\hline
    \end{tabular}
    \label{tab:dynamic}
    \end{table*}
\subsubsection{Static HDR Scenes}
Evaluations are first carried out on the synthetic dataset, \ie\ \mdata-S, and real-world static dataset, \ie\ \mdata-Rsimple, where the ground-truth sharp HDR images are available, and we provide the timestamps of the ground-truth images for all the compared methods to satisfy them generating sharp HDR images at corresponding timestamps. Comparisons with state-of-the-art methods are conducted both quantitatively and qualitatively.

\noindent\textbf{Quantitative Evaluation.}
Quantitative results in \cref{tab:EBL2SH-S} show that our method performs best in all metrics, demonstrating our algorithm's ability to handle hybrid degradation of LDR and global blur. Both HDRI+Deblurring and Deblurring+HDRI achieve comparable results to our method on \mdata-S, with only a 1.94/0.042/2.15 drop on PSNR-$\mu$/SSIM-$\mu$/HDR-VDP-2, signifying a commendable restoration of image content. However, in the real-world dataset \mdata-Rsimple, the hybrid degradation is more severely coupled, and the degenerate decoupling process is more complex, leading to the deteriorated performance of these cascading methods. Thanks to our end-to-end framework for joint HDRI and deblurring, this problem can be resolved by learning brightness variation from events in both temporal and spatial domains. As shown in \cref{tab:EBL2SH-S}, our proposed \mname\ performs favorably against the state-of-the-art methods, especially a large margin on PSNR-$\mu$ of 6.43 dB and HDR-VDP-2 of 3.70. Moreover, the comparison in terms of runtime demonstrates the efficiency of our method over the compared methods.

\noindent\textbf{Qualitative Evaluation.}
We present the visual comparison of results on two typical blurry LDR images in \cref{fig:static} and analyze them regarding HDR recovery, motion deblurring, and visual quality. 

Even though HDRI+Deblurring and Deblurring+HDRI both struggle to reconstruct sharp HDR images, they still can not recover sufficient details due to the complex hybrid degradation. When performing HDRI+Deblurring, the outputs lack detailed structures in the saturated regions because the presence of blur aggravates the performance drop of HDRI methods. Thus, the following deblurring can still generate precise results, \eg the details in the red box of the first example are corrupted. Besides, the results of Deblurring+HDRI methods suffer from blur in the saturated regions because existing deblurring methods only leverage events to handle motion blur while incorrectly estimating the motion kernel in the region which is badly exposed, leading to the decreased performance of followed HDRI process, \eg the light in the green box of the first example and the structures in color boxes of the second example still contain blur. Furthermore, the results of cascade methods also suffer from color distortion and unnatural textures.

By comparison, our \mname\ leverages both the high dynamic range and high temporal resolution of events and conducts self-supervised losses to jointly optimize our framework, thus generating more visually pleasing, informative, and sharp clear HDR images that exhibit the most similar appearances to the ground-truth images.

\subsubsection{Dynamic HDR Scenes}
The E-BL2SH task in real-world dynamic HDR scenes is much more challenging because images suffer from both global partial blur and saturated brightness, resulting in more severe edge degradation of moving objects. Since ground-truth sharp HDR images in dynamic scenes are difficult to obtain, we compare our \mname\ to the cascading methods quantitatively on \mdata-Rcomplex with a series of no-reference metrics and qualitatively analyze the output results from different perspectives.

\noindent\textbf{Quantitative Evaluation.}
We first offer the overall comparison in \cref{tab:dynamic}, and then provide an in-depth analysis of each metric. It can be seen that our proposed \mname\ achieves superior performance in terms of all metrics from diverse perspectives, which verifies the effectiveness. Specifically, our proposed method achieves the best value in AG, indicating that our results contain richer details and edge information. In addition, when it comes to image quality, our \mname\ brings remarkable improvements compared to the state-of-the-art methods, on average 0.203 gain on NIQE, 0.111 on CLIP-IQA+, 0.100 on MANIQA, 0.103 on MUSIQ, and 1.189 on LIQE. 

\begin{figure}[!t]
\centering
\includegraphics[width=1\columnwidth]{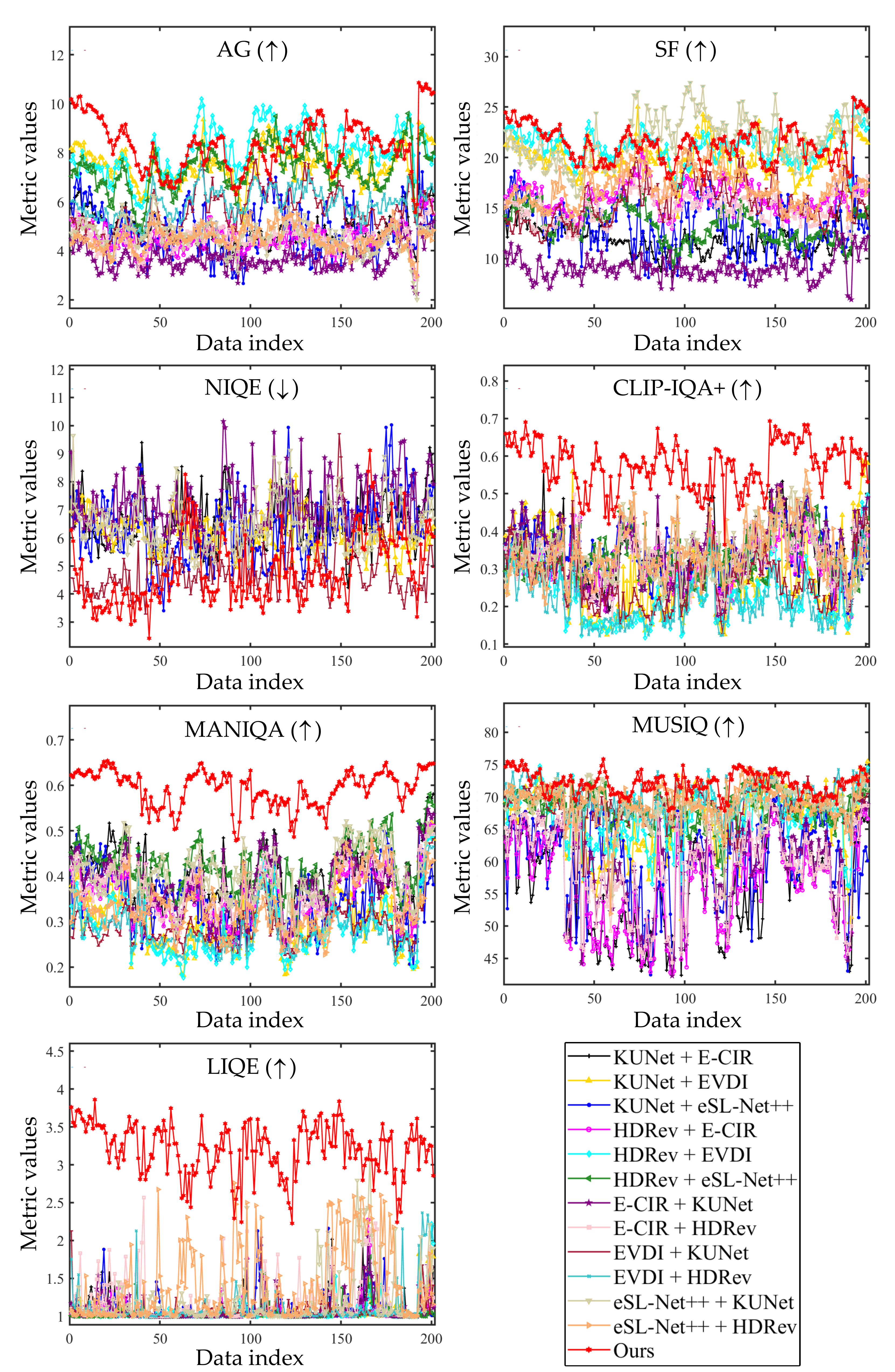} 
\caption{Per-image quantitative comparisons in terms of different metrics.}
\label{fig:real_each}
\end{figure}

To reveal the robustness and wide applicability of our method, we randomly select 200 data pairs from the test results and provide a detailed per-image comparison on different metrics in \cref{fig:real_each}. We can see that our method reaches a stable and satisfactory performance on different images, particularly on the state-of-the-art image quality assessment metrics such as CLIP-IQA+, MANIQA, and LIQE, where each of our results outperforms the comparative methods. Although cascading HDRev and EVDI obtains comparable results on AG and SF, the image quality of the results obtained is unsatisfactory.

\begin{figure*}[!t]
\centering
\includegraphics[width=1\textwidth]{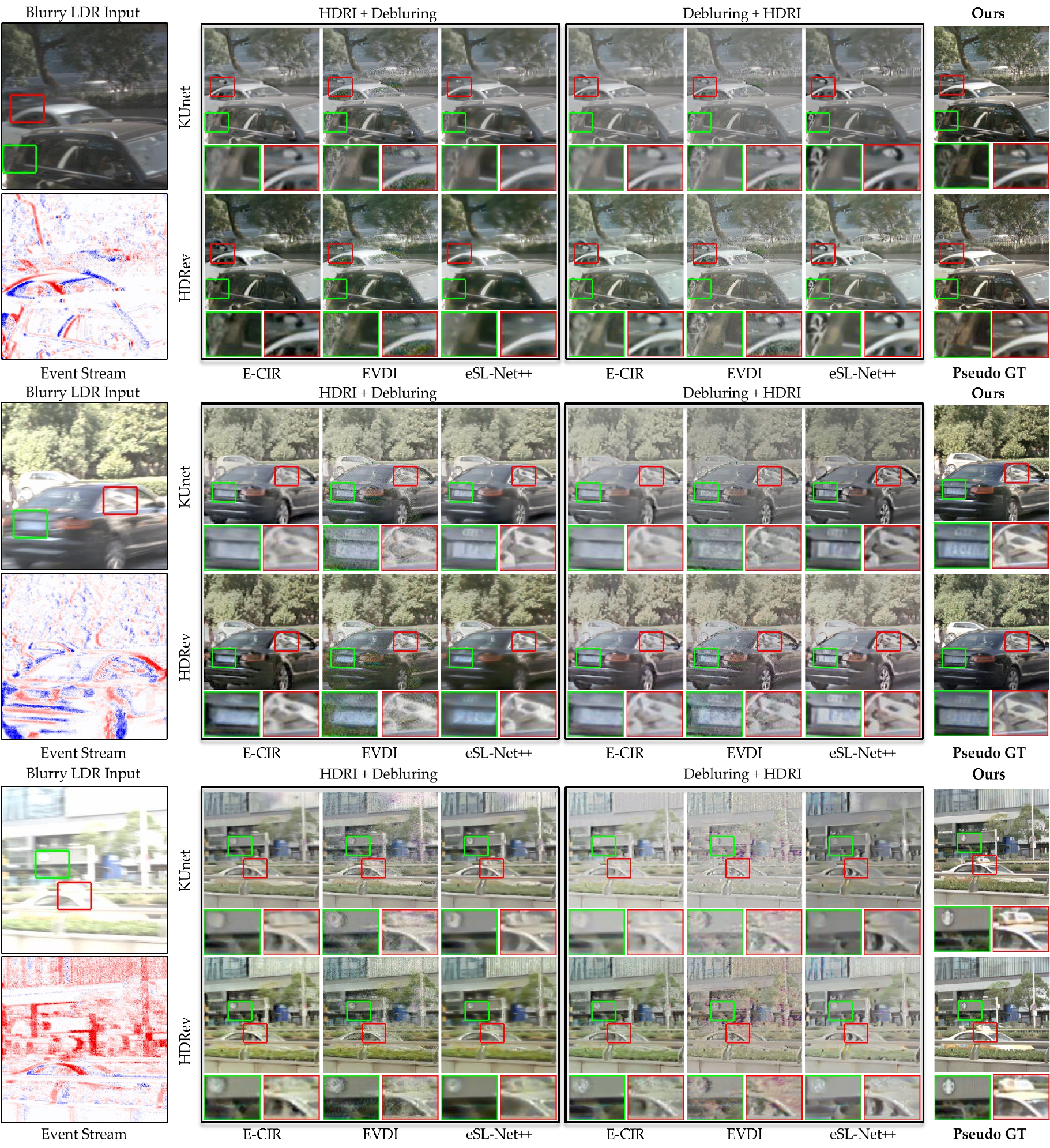} 
\caption{Qualitative results on the real-world dynamic dataset \mdata-Rcomplex, which are tone-mapped by using the method proposed by Reinhard \etal~\cite{reinhard2002photographic}. The pseudo GT is generated by globally aligning and merging three alternately exposed images near the reconstructed timestamp, which may contain ghosts due to fast-moving targets. We present it for the visual reference of image content and color.}
\label{fig:real_world_dynamic}
\end{figure*}

\begin{figure*}[htbp]
\centering
\includegraphics[width=1\textwidth]{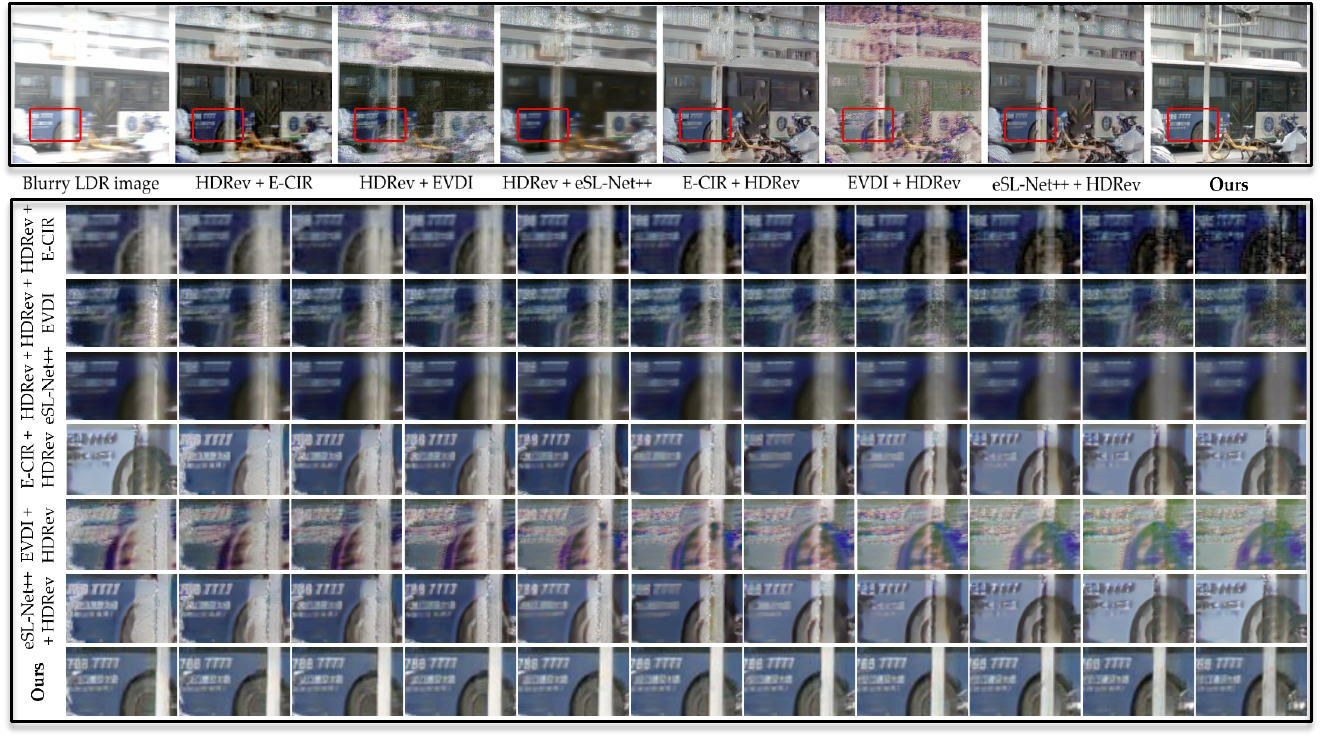} 
\caption{Qualitative results of sharp HDR sequence reconstructed from a blurry LDR image by E-BL2SH methods.}
\label{fig:real_world_sequence}
\end{figure*}

\noindent\textbf{Qualitative Evaluation.}
For qualitative evaluation, we provide the pseudo GT generated by globally aligning and merging three alternately exposed images as the reference and present visual comparisons on several typical images captured under different light conditions in \cref{fig:real_world_dynamic}. As can be seen from the results, although the input images are severely degraded by badly exposed regions and non-linear blur, our results can still maintain promising detailed structures and are reconstructed to a high dynamic range as well as the pseudo GT. By contrast, other comparison methods, especially Deblurring+HDRI ones, lose information on structure, dynamic range, as well as colors. For example, in \cref{fig:real_world_dynamic}, cascading methods suffer from noticeable noise, low contrast, and color artifacts when processing blurry under-exposed regions. Additionally, the results also show brightness transition artifacts when dealing with blurry over-exposed regions, \eg the car in the red box of the third example still undergoes hybrid degradation and lack of details. In our concern, the unsatisfactory appearance of the compared methods is mainly caused by three reasons: (1) The sub-optimal outcome resulting from the cascading strategy. For either HDRI+Deblurring methods or Deblurring+HDRI methods, simply cascading two networks may yield sub-optimal solutions, because the quality of final outputs is contingent upon the performance of the initial stage; (2) The loss of brightness consistency under abnormal exposure. For Deblurring+HDRI methods which take deblurring methods as the initial stage, processing deblurring on the abnormal exposed regions is exertive because saturated pixels hide the accurate information of brightness, resulting in the loss of brightness consistency prior that deblurring methods rely on; (3) The loss of structural prior in blurry regions. Since HDRI+Deblurring methods take HDRI methods as the pre-process, the restoration of dynamic range is particularly important for subsequent deblurring. However, current HDRI methods cannot handle blurry LDR images because the structural information they rely on to restore saturated regions is hidden by blur.

To further demonstrate our performance in recovering sharp HDR sequences, we show more results in \cref{fig:real_world_sequence}, where the blurry LDR image is reconstructed to 11 frames. The event-based methods ranked high in \cref{tab:dynamic} are employed for comparison. It is obvious that their reconstructed results still exhibit blur artifacts, saturated regions, and color artifacts, particularly when the EVDI technique is employed as the deblurring process. This occurs because the majority of self-supervised constraints in EVDI strictly rely on event-based double integration, which is invalid when the HDRI methods cannot recover all the saturated regions, as illustrated in the 2nd row and 5th row of \cref{fig:real_world_sequence}. Benefiting from the elaborately designed \mnet\ network and self-supervised training strategy, our method successfully reconstructs the intermediate sharp HDR frames from the severely blurry LDR image captured in dynamic HDR scenes with arbitrary complex motion, \eg the restoration from the fast-moving car that is overexposed and its spinning wheel that is underexposed.

\begin{table*}[htbp]
\caption{Quantitative comparison of object detection on dynamic HDR scenes with motion blur. Best results are shown \textbf{bold}, the second best as \underline{underline}.}\label{yolo}
\begin{center}	
\begin{tabular}{ccccccccccc}
\hline
\multirow{2}{*}{Detectors} &\multirow{2}{*}{Metrics} & \multirow{2}{*}{Input} & HDRev & HDRev & HDRev & E-CIR & EVDI & eSL-Net++ &\multirow{2}{*}{Ours} \\
& & & E-CIR & EVDI & eSL-Net++ & HDRev & HDRev & HDRev & \\ 
\hline
\multirow{2}{*}{YOLOV5s} & mAP50 & 0.477 & 0.474 & 0.279 & 0.465 & \underline{0.496} & 0.227 & 0.486 & \textbf{0.664} \\  
& mAP50-90 & 0.291 & 0.283 & 0.156 & 0.283 & \underline{0.312} & 0.122 & 0.293 & \textbf{0.469} \\
\hline
\multirow{2}{*}{YOLOV8x} & mAP50 & \underline{0.646}  & 0.505 & 0.418 & 0.493 & 0.521 & 0.304 & 0.507 & \textbf{0.830} \\
& mAP50-90 & \underline{0.446}  & 0.332 & 0.270 & 0.344 & 0.350 & 0.185 & 0.343& \textbf{0.722} \\
\hline
\end{tabular}
\end{center}
\end{table*}

\begin{figure}[!t]
\centering
\includegraphics[width=1\columnwidth]{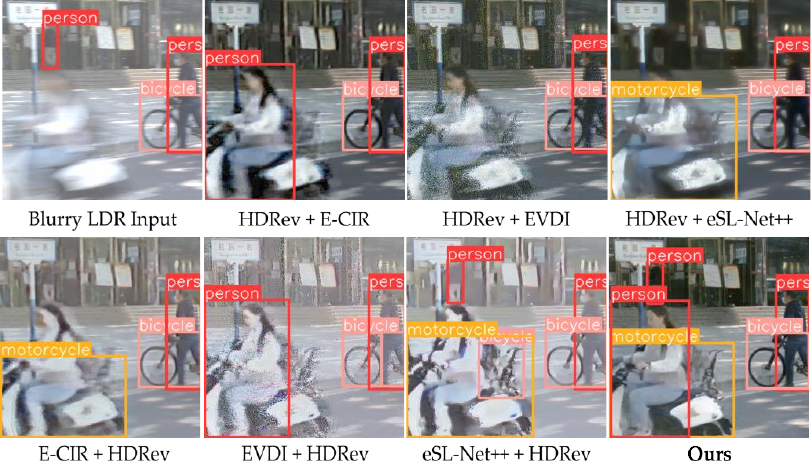} 
\caption{Detection results of YOLOv8x on blurry LDR images preprocessed by different E-BL2SH methods.}
\label{fig:yolopic}
\end{figure}

\begin{table*}[!t]
\centering
\caption{Comparison on the real-world dataset of \mdata\ in ablation settings of network structures. We highlight the best performance as \textbf{bold}, the second best as \underline{underline}.} \label{tab:abnet}
\begin{tabular}{cccc|ccc|ccc}
\hline
\multicolumn{4}{c|}{Ablation Setting} & \multicolumn{3}{c|}{\mdata-Rsimple} & \multicolumn{3}{c}{\mdata-Rcomplex} \\
\hline
MRF & Residual & DRC & Pre-train &PSNR-$\mu$ & SSIM-$\mu$ & HDR-VDP-2& CLIP-IQA+ & MANIQA & MUSIQ\\
\hline
 & \checkmark& \checkmark& \checkmark & 24.69 & 0.846 & 45.84& 0.485 &0.586& 70.628\\
\checkmark & & \checkmark &\checkmark& 24.74 & 0.850 & 46.44& 0.454&0.581 & 70.745\\
\checkmark & \checkmark &  & \checkmark & 23.89 & 0.843 & 45.29 & 0.468& 0.581 & 71.470\\
\checkmark & \checkmark & \checkmark & & \underline{26.77} & \underline{0.866} & \underline{48.63}& \underline{0.487} &\underline{0.592}& \underline{71.847} \\
\checkmark & \checkmark & \checkmark & \checkmark & \textbf{27.36} & \textbf{0.880} & \textbf{49.05} & \textbf{0.518} & \textbf{0.606} & \textbf{71.899}\\
\hline
\end{tabular}
\end{table*}

\begin{figure*}[!t]
\centering
\includegraphics[width=0.90\textwidth]{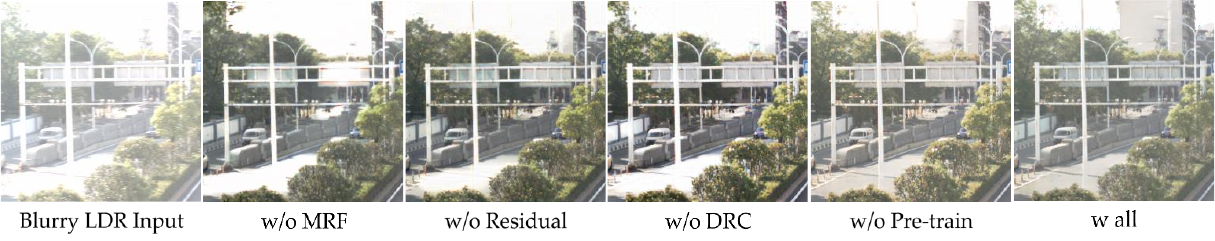} 
\caption{Visual comparison on the real-world dataset of \mdata\ in ablation settings of network structures.}
\label{fig:abnet}
\end{figure*}

\subsection{External Verification on Object Detection}
\label{Object detection}
To evaluate the practical benefits of our \mname\ and its improved performance, external verification is further performed. We take the E-BL2SH methods as a pre-processing step for object detection and compare the detection results on 796 typical images in dynamic HDR scenes by using YOLOV5s or YOLOV8x as the detector~\cite{jocher2023yolo}. The mean Average Precision (mAP) shown in \cref{yolo} proves that our method is able to improve the accuracy of object detection on blurry LDR images. Specifically, most E-BL2SH techniques improve the performance of YOLOV5s, and our \mname\ outperforms the compared methods by 0.168 in mAP50 and 0.157 in mAP50-90. However, when employing the more efficient detector, YOLOV8x, none of compared E-BL2SH methods are able to yield positive gains, whereas our proposed method still benefits the detection with a notable improvement of 28.4\% in mAP50 and 61.9\% in mAP50-90. \cref{fig:yolopic} shows the visual detection results of YOLOV8x by taking E-BL2SH methods as the pre-process, we can observe that more objects are detected in our result, especially objects with hybrid degradation.



\subsection{Ablation Study}
\label{Ablation Study}
The effectiveness of the framework is evaluated in this subsection, including architectures and optimization strategies.

\subsubsection{Analysis of Network Architecture}
We conduct ablation studies on two key modules in our framework, \ie the MRFR block and the DRC module, and present the experimental results on the real-world dataset in \cref{tab:abnet} and \cref{fig:abnet}.

\noindent\textbf{Importance of MRFR Module.} Since the valuable information in blurry LDR images and events is regionally distributed, the MRFR block is essential to extract multiscale features and establish interregional connections. Therefore, two experiments are conducted to verify the effectiveness of the Multiple Receptive Field (MRF) and residual structures. In our experimental setup, we disable the MRF structure by replacing dense multiscale blocks with dense blocks $3\times3$ and disable the residual structure by removing the residual connections between different dense blocks. As illustrated in \cref{fig:abnet}, integrating the MRF and residual structures in \mname\ effectively harnesses events' high dynamic range and temporal resolution, resulting in superior reconstruction results. Their utilization also yields remarkable enhancements in terms of all the real-world dataset metrics, including static and dynamic scenes.

\noindent\textbf{Importance of DRC Module.}
We further validate the DRC module, focusing on its contribution and the effectiveness of the pre-training strategy. In the first experiment, we remove the DRC module and directly measure the difference between $\hat{I}(t)$ and $\tilde{I}(t)$. This configuration, called without (w/o) DRC, yields inferior results, as indicated in \cref{tab:abnet}, with decreases of 3.47/0.037/3.76 in PSNR-$\mu$/SSIM-$\mu$/HDR-VDP-2 on static scenes and 0.050/0.025/0.429 in CLIP-IQA+/MANIQA/MUSIQ in dynamic scenes. In the second experiment, the DRC module is optimized without the pre-training that is described in \cref{Implementation Details}, denoted as w/o pre-train. The results in \cref{tab:abnet} indicate that the pre-training strategy substantially improves performance, as evidenced by the significant improvement in all evaluation metrics. This enhancement is also visually proven in \cref{fig:abnet}, where the DRC module and pre-training strategy effectively preserve the details in over-exposed regions.

\subsubsection{Analysis of Optimization Strategy}
\begin{table*}[bpht]
\centering
\caption{Comparison on the real-world dataset of \mdata\ in ablation settings of loss functions. We highlight the best performance as \textbf{bold}, the second best as \underline{underline}.} \label{abltable}
\begin{tabular}{cccc|ccc|ccc}
\hline
\multicolumn{4}{c|}{Ablation Setting} & \multicolumn{3}{c|}{\mdata-Rsimple} & \multicolumn{3}{c}{\mdata-Rcomplex} \\
\hline
$\mathcal{L}_{HL}$ & $\mathcal{L}_{LL}$ & $\mathcal{L}_{LH}$ & $\mathcal{L}_{HH}$ &PSNR-$\mu$ & SSIM-$\mu$ & HDR-VDP-2& CLIP-IQA+ & MANIQA & MUSIQ \\
\hline
           & \checkmark & \checkmark & \checkmark & 10.72 & 0.710 & 30.56& \underline{0.515} & 0.453& 52.322 \\
\checkmark &            & \checkmark & \checkmark & \underline{22.79} & \underline{0.837} & 42.73& 0.467& 0.569& 70.136 \\
\checkmark & \checkmark &            & \checkmark & 13.48 & 0.711 & 43.98& 0.135 & 0.367& 70.761 \\
\checkmark & \checkmark & \checkmark &            & 19.10 & 0.807 & \underline{46.20} & 0.391 &\underline{0.595} & \textbf{74.952} \\
\checkmark & \checkmark & \checkmark & \checkmark & \textbf{27.36} & \textbf{0.880} & \textbf{49.05} & \textbf{0.518} & \textbf{0.606} & \underline{71.899} \\
\hline
\end{tabular}
\end{table*}

\begin{figure*}[!htb]
\centering
\includegraphics[width=0.9\textwidth]{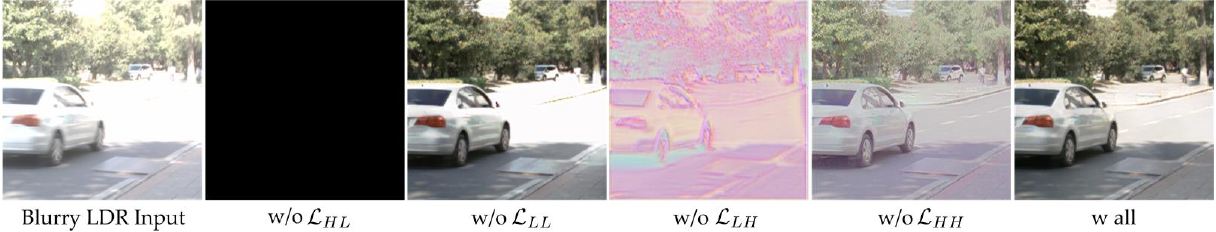} 
\caption{Visual comparison on the real-world dataset of \mdata\ in ablation settings of loss functions.}
\label{fig:abpic}
\end{figure*}

To investigate the contribution of each loss function on the overall framework and demonstrate their interplay relations, we execute comprehensive ablation studies on the real-world dataset and present the quantitative results in \cref{abltable}. Meanwhile, the corresponding qualitative ablation results are shown in \cref{fig:abpic}. 

\noindent\textbf{Importance of \bm{$\mathcal{L}_{HL}$}.} In our \mname, the decomposition consistency loss $\mathcal{L}_{HL}$ plays the most important role since it establishes the relationship between the blurry LDR image, events, and sharp LDR images, which facilitates events to provide reliable dynamic range and high temporal resolution. The visualization of the reconstructed sharp HDR image, as illustrated in \cref{fig:abpic}, reveals the absence of informative content when $\mathcal{L}_{HL}$ is omitted, substantiating the importance of $\mathcal{L}_{HL}$. Correspondingly, the quantitative evaluation in \cref{abltable} corroborates this observation, as it showcases the worst results when $\mathcal{L}_{HL}$ is excluded.

\noindent\textbf{Importance of \bm{$\mathcal{L}_{LL}$}.}
Although $\mathcal{L}_{HL}$ provides meaningful supervision for the DRD module, it can not conduct complete constrain on the decomposed LDR stack to keep the balance in HDR-to-LDR transformation without $\mathcal{L}_{LL}$, \ie the over-exposed car and road can not be recovered in \cref{fig:abpic}. By introducing $\mathcal{L}_{LL}$, the brightness of the decomposed images is maintained at different exposure levels; therefore, \mname\ can recover more details in HDR scenes. As shown in \cref{abltable}, the utilization of $\mathcal{L}_{LL}$ achieves the PSNR-$\mu$/SSIM-$\mu$/HDR-VDP-2 gain up to 4.57/0.043/6.32 on static scenes and CLIP-IQA+/MANIQA/MUSIQ gain up to 0.033/0.020/1.271 on dynamic scenes. Qualitatively, $\mathcal{L}_{LL}$ significantly improves the dynamic range, leading to more informative outputs.

\noindent\textbf{Importance of \bm{$\mathcal{L}_{LH}$}.} Since there are no usable sharp HDR ground truth images to guide the optimization of the dynamic range fusion process, the fusion consistency loss $\mathcal{L}_{LH}$ is introduced to constrain the DRC module, encouraging the network to reconstruct a visually pleasant sharp HDR image from bracketed sharp LDR images. As shown in \cref{fig:abpic}, introducing the fusion consistency loss leads to accurate and visually appealing predicted results. Notably, as demonstrated in \cref{abltable}, the commissioning effect of $\mathcal{L}_{LH}$ on the network is surpassed only by $\mathcal{L}_{HL}$, highlighting its importance in the optimization process.

\noindent\textbf{Importance of \bm{$\mathcal{L}_{HH}$}.} The reconstruction consistency loss $\mathcal{L}_{HH}$ conducts self-supervised consistency between the main branch and the assistance branch, guaranteeing the output of the main branch is similar to that of the assistance one. Even though the results contain enough details when training without $\mathcal{L}_{HH}$, they tend to suffer from lower contrast, as shown in \cref{fig:abpic}. Meanwhile, results in \cref{abltable} show that the supervision of $\mathcal{L}_{HH}$ obtains significant performance improvement, \ie 8.26 of PSNR-$\mu$, 0.073 of SSIM-$\mu$, 2.85 of HDR-VDP-2, 0.127 of CLIQ-IQA+ and 0.011 of MANIQA.

\section{Conclusion}
\label{Conclusion}
In this paper, we propose a self-supervised event-based HDRI method, \ie\ \mname\ to reconstruct the latent sharp HDR sequence from a blurry LDR image and its concurrent event streams. Specifically, a novel self-supervised framework is proposed to perform optimization on sharp LDR observations by leveraging the dynamic range decomposition and composition operations, which addresses the problem that ground-truth sharp HDR images are difficult to obtain in highly dynamic scenes. Then, we introduce an end-to-end network for joint HDRI and motion deblurring, which efficiently handles the hybrid degradation by leveraging the high dynamic range and high temporal resolution of events simultaneously. We also construct a new HDRI dataset, \ie\ \mdata, which contains aligned blurry LDR images, event streams, and concurrent sharp LDR frames. Quantitative and qualitative results over the dataset show the effectiveness of our proposed \mname\ method.



%





\ifCLASSOPTIONcaptionsoff
  \newpage
\fi



\bibliographystyle{IEEEtran}
\bibliography{bare_jrnl_compsoc}

\begin{thebibliography}{10}
\providecommand{\url}[1]{#1}
\csname url@samestyle\endcsname
\providecommand{\newblock}{\relax}
\providecommand{\bibinfo}[2]{#2}
\providecommand{\BIBentrySTDinterwordspacing}{\spaceskip=0pt\relax}
\providecommand{\BIBentryALTinterwordstretchfactor}{4}
\providecommand{\BIBentryALTinterwordspacing}{\spaceskip=\fontdimen2\font plus
\BIBentryALTinterwordstretchfactor\fontdimen3\font minus
  \fontdimen4\font\relax}
\providecommand{\BIBforeignlanguage}[2]{{%
\expandafter\ifx\csname l@#1\endcsname\relax
\typeout{** WARNING: IEEEtran.bst: No hyphenation pattern has been}%
\typeout{** loaded for the language `#1'. Using the pattern for}%
\typeout{** the default language instead.}%
\else
\language=\csname l@#1\endcsname
\fi
#2}}
\providecommand{\BIBdecl}{\relax}
\BIBdecl

\bibitem{wang2021deep}
L.~Wang and K.-J. Yoon, ``Deep learning for hdr imaging: State-of-the-art and
  future trends,'' \emph{IEEE Trans. Pattern Anal. Mach. Intell.}, vol.~44,
  no.~12, pp. 8874--8895, 2021.

\bibitem{lu2009high}
P.-Y. Lu, T.-H. Huang, M.-S. Wu, Y.-T. Cheng, and Y.-Y. Chuang, ``High dynamic
  range image reconstruction from hand-held cameras,'' in \emph{IEEE Conf.
  Comput. Vis. Pattern Recog.}\hskip 1em plus 0.5em minus 0.4em\relax IEEE,
  2009, pp. 509--516.

\bibitem{chen2020learning}
Y.~Chen, G.~Jiang, M.~Yu, Y.~Yang, and Y.-S. Ho, ``Learning stereo high dynamic
  range imaging from a pair of cameras with different exposure parameters,''
  \emph{IEEE Trans. on Computational Imaging}, vol.~6, pp. 1044--1058, 2020.

\bibitem{chen2021hdrunet}
X.~Chen, Y.~Liu, Z.~Zhang, Y.~Qiao, and C.~Dong, ``Hdrunet: Single image hdr
  reconstruction with denoising and dequantization,'' in \emph{IEEE Conf.
  Comput. Vis. Pattern Recog.}, 2021, pp. 354--363.

\bibitem{wang2022kunet}
H.~Wang, M.~Ye, X.~Zhu, S.~Li, C.~Zhu, and X.~Li, ``Kunet: Imaging
  knowledge-inspired single hdr image reconstruction,'' in \emph{IJCAI}, 2022.

\bibitem{kalantari2017deep}
N.~K. Kalantari, R.~Ramamoorthi \emph{et~al.}, ``Deep high dynamic range
  imaging of dynamic scenes.'' \emph{ACM Transactions on Graphics.}, vol.~36,
  no.~4, pp. 144--1, 2017.

\bibitem{xu2020fusiondn}
H.~Xu, J.~Ma, Z.~Le, J.~Jiang, and X.~Guo, ``Fusiondn: A unified densely
  connected network for image fusion,'' in \emph{Proceedings of the AAAI
  conference on artificial intelligence}, vol.~34, no.~07, 2020, pp.
  12\,484--12\,491.

\bibitem{xu2020mef}
H.~Xu, J.~Ma, and X.-P. Zhang, ``Mef-gan: Multi-exposure image fusion via
  generative adversarial networks,'' \emph{IEEE Trans. Image Process.},
  vol.~29, pp. 7203--7216, 2020.

\bibitem{yang2023learning}
Y.~Yang, J.~Han, J.~Liang, I.~Sato, and B.~Shi, ``Learning event guided high
  dynamic range video reconstruction,'' in \emph{IEEE Conf. Comput. Vis.
  Pattern Recog.}, 2023, pp. 13\,924--13\,934.

\bibitem{han2020neuromorphic}
J.~Han, C.~Zhou, P.~Duan, Y.~Tang, C.~Xu, C.~Xu, T.~Huang, and B.~Shi,
  ``Neuromorphic camera guided high dynamic range imaging,'' in \emph{IEEE
  Conf. Comput. Vis. Pattern Recog.}, 2020, pp. 1730--1739.

\bibitem{messikommer2022multi}
N.~Messikommer, S.~Georgoulis, D.~Gehrig, S.~Tulyakov, J.~Erbach,
  A.~Bochicchio, Y.~Li, and D.~Scaramuzza, ``Multi-bracket high dynamic range
  imaging with event cameras,'' in \emph{IEEE Conf. Comput. Vis. Pattern
  Recog.}, 2022, pp. 547--557.

\bibitem{purohit2020region}
K.~Purohit and A.~Rajagopalan, ``Region-adaptive dense network for efficient
  motion deblurring,'' in \emph{Proceedings of the AAAI conference on
  artificial intelligence}, vol.~34, no.~07, 2020, pp. 11\,882--11\,889.

\bibitem{purohit2019bringing}
K.~Purohit, A.~Shah, and A.~Rajagopalan, ``Bringing alive blurred moments,'' in
  \emph{IEEE Conf. Comput. Vis. Pattern Recog.}, 2019, pp. 6830--6839.

\bibitem{haoyu2020learning}
C.~Haoyu, T.~Minggui, S.~Boxin, W.~YIzhou, and H.~Tiejun, ``Learning to deblur
  and generate high frame rate video with an event camera,'' \emph{arXiv
  preprint arXiv:2003.00847}, 2020.

\bibitem{xu2021motion}
F.~Xu, L.~Yu, B.~Wang, W.~Yang, G.-S. Xia, X.~Jia, Z.~Qiao, and J.~Liu,
  ``Motion deblurring with real events,'' in \emph{Int. Conf. Comput. Vis.},
  2021, pp. 2583--2592.

\bibitem{song2022cir}
C.~Song, Q.~Huang, and C.~Bajaj, ``E-cir: Event-enhanced continuous intensity
  recovery,'' in \emph{IEEE Conf. Comput. Vis. Pattern Recog.}, 2022, pp.
  7803--7812.

\bibitem{gallego2020event}
G.~Gallego, T.~Delbr{\"u}ck, G.~Orchard, C.~Bartolozzi, B.~Taba, A.~Censi,
  S.~Leutenegger, A.~J. Davison, J.~Conradt, K.~Daniilidis \emph{et~al.},
  ``Event-based vision: A survey,'' \emph{IEEE Trans. Pattern Anal. Mach.
  Intell.}, vol.~44, no.~1, pp. 154--180, 2020.

\bibitem{zhang2021event}
X.~Zhang, W.~Liao, L.~Yu, W.~Yang, and G.-S. Xia, ``Event-based synthetic
  aperture imaging with a hybrid network,'' in \emph{IEEE Conf. Comput. Vis.
  Pattern Recog.}, 2021, pp. 14\,235--14\,244.

\bibitem{chen2021hdr}
G.~Chen, C.~Chen, S.~Guo, Z.~Liang, K.-Y.~K. Wong, and L.~Zhang, ``Hdr video
  reconstruction: A coarse-to-fine network and a real-world benchmark
  dataset,'' in \emph{Int. Conf. Comput. Vis.}, 2021, pp. 2502--2511.

\bibitem{robidoux2021end}
N.~Robidoux, L.~E.~G. Capel, D.-e. Seo, A.~Sharma, F.~Ariza, and F.~Heide,
  ``End-to-end high dynamic range camera pipeline optimization,'' in \emph{IEEE
  Conf. Comput. Vis. Pattern Recog.}, 2021, pp. 6297--6307.

\bibitem{onzon2021neural}
E.~Onzon, F.~Mannan, and F.~Heide, ``Neural auto-exposure for high-dynamic
  range object detection,'' in \emph{IEEE Conf. Comput. Vis. Pattern Recog.},
  2021, pp. 7710--7720.

\bibitem{banterle2017advanced}
F.~Banterle, A.~Artusi, K.~Debattista, and A.~Chalmers, \emph{Advanced high
  dynamic range imaging}.\hskip 1em plus 0.5em minus 0.4em\relax AK Peters/CRC
  Press, 2017.

\bibitem{eilertsen2017hdr}
G.~Eilertsen, J.~Kronander, G.~Denes, R.~K. Mantiuk, and J.~Unger, ``Hdr image
  reconstruction from a single exposure using deep cnns,'' \emph{ACM
  Transactions on Graphics.}, vol.~36, no.~6, pp. 1--15, 2017.

\bibitem{marnerides2018expandnet}
D.~Marnerides, T.~Bashford-Rogers, J.~Hatchett, and K.~Debattista, ``Expandnet:
  A deep convolutional neural network for high dynamic range expansion from low
  dynamic range content,'' in \emph{Computer Graphics Forum}, vol.~37,
  no.~2.\hskip 1em plus 0.5em minus 0.4em\relax Wiley Online Library, 2018, pp.
  37--49.

\bibitem{debevec1997recovering}
P.~Debevec and J.~Malik, ``Recovering high dynamic range radiance maps from
  photographs: Proceedings of the 24th annual conference on computer graphics
  and interactive techniques,'' \emph{Los Angeles, USA: SIGGRAPH}, 1997.

\bibitem{nazarczuk2022self}
M.~Nazarczuk, S.~Catley-Chandar, A.~Leonardis, and E.~P{\'e}rez-Pellitero,
  ``Self-supervised hdr imaging from motion and exposure cues,'' \emph{arXiv
  preprint arXiv:2203.12311}, 2022.

\bibitem{han2023hybrid}
J.~Han, Y.~Yang, P.~Duan, C.~Zhou, L.~Ma, C.~Xu, T.~Huang, I.~Sato, and B.~Shi,
  ``Hybrid high dynamic range imaging fusing neuromorphic and conventional
  images,'' \emph{IEEE Trans. Pattern Anal. Mach. Intell.}, 2023.

\bibitem{zhu2019deformable}
X.~Zhu, H.~Hu, S.~Lin, and J.~Dai, ``Deformable convnets v2: More deformable,
  better results,'' in \emph{IEEE Conf. Comput. Vis. Pattern Recog.}, 2019, pp.
  9308--9316.

\bibitem{pan2020high}
L.~Pan, R.~Hartley, C.~Scheerlinck, M.~Liu, X.~Yu, and Y.~Dai, ``High frame
  rate video reconstruction based on an event camera,'' \emph{IEEE Trans.
  Pattern Anal. Mach. Intell.}, 2020.

\bibitem{zhang2022unifying}
X.~Zhang and L.~Yu, ``Unifying motion deblurring and frame interpolation with
  events,'' in \emph{IEEE Conf. Comput. Vis. Pattern Recog.}, 2022, pp.
  17\,765--17\,774.

\bibitem{zhou2023deblurring}
C.~Zhou, M.~Teng, J.~Han, J.~Liang, C.~Xu, G.~Cao, and B.~Shi, ``Deblurring
  low-light images with events,'' \emph{Int. J. Comput. Vis.}, vol. 131, no.~5,
  pp. 1284--1298, 2023.

\bibitem{wang2020event}
B.~Wang, J.~He, L.~Yu, G.-S. Xia, and W.~Yang, ``Event enhanced high-quality
  image recovery,'' in \emph{Eur. Conf. Comput. Vis.}\hskip 1em plus 0.5em
  minus 0.4em\relax Springer, 2020, pp. 155--171.

\bibitem{yu2023learning}
L.~Yu, B.~Wang, X.~Zhang, H.~Zhang, W.~Yang, J.~Liu, and G.-S. Xia, ``Learning
  to super-resolve blurry images with events,'' \emph{IEEE Trans. Pattern Anal.
  Mach. Intell.}, 2023.

\bibitem{vasu2018joint}
S.~Vasu, A.~Shenoi, and A.~Rajagopazan, ``Joint hdr and super-resolution
  imaging in motion blur,'' in \emph{IEEE Int. Conf. Image Process.}\hskip 1em
  plus 0.5em minus 0.4em\relax IEEE, 2018, pp. 2885--2889.

\bibitem{kim2019deep}
S.~Y. Kim, J.~Oh, and M.~Kim, ``Deep sr-itm: Joint learning of super-resolution
  and inverse tone-mapping for 4k uhd hdr applications,'' in \emph{Int. Conf.
  Comput. Vis.}, 2019, pp. 3116--3125.

\bibitem{deng2021deep}
X.~Deng, Y.~Zhang, M.~Xu, S.~Gu, and Y.~Duan, ``Deep coupled feedback network
  for joint exposure fusion and image super-resolution,'' \emph{IEEE Trans.
  Image Process.}, vol.~30, pp. 3098--3112, 2021.

\bibitem{perez2021ntire}
E.~P{\'e}rez-Pellitero, S.~Catley-Chandar, A.~Leonardis, and R.~Timofte,
  ``Ntire 2021 challenge on high dynamic range imaging: Dataset, methods and
  results,'' in \emph{IEEE Conf. Comput. Vis. Pattern Recog.}, 2021, pp.
  691--700.

\bibitem{akyuz2020deep}
A.~O. Aky{\"u}z \emph{et~al.}, ``Deep joint deinterlacing and denoising for
  single shot dual-iso hdr reconstruction,'' \emph{IEEE Trans. Image Process.},
  vol.~29, pp. 7511--7524, 2020.

\bibitem{kim2019multi}
S.~Y. Kim and M.~Kim, ``A multi-purpose convolutional neural network for
  simultaneous super-resolution and high dynamic range image reconstruction,''
  in \emph{ACCV}.\hskip 1em plus 0.5em minus 0.4em\relax Springer, 2019, pp.
  379--394.

\bibitem{johnson2016perceptual}
J.~Johnson, A.~Alahi, and L.~Fei-Fei, ``Perceptual losses for real-time style
  transfer and super-resolution,'' in \emph{Eur. Conf. Comput. Vis.}\hskip 1em
  plus 0.5em minus 0.4em\relax Springer, 2016, pp. 694--711.

\bibitem{wan2022old}
Z.~Wan, B.~Zhang, D.~Chen, P.~Zhang, F.~Wen, and J.~Liao, ``Old photo
  restoration via deep latent space translation,'' \emph{IEEE Transactions on
  Pattern Analysis and Machine Intelligence}, vol.~45, no.~2, pp. 2071--2087,
  2022.

\bibitem{dong2020multi}
H.~Dong, J.~Pan, L.~Xiang, Z.~Hu, X.~Zhang, F.~Wang, and M.-H. Yang,
  ``Multi-scale boosted dehazing network with dense feature fusion,'' in
  \emph{IEEE Conf. Comput. Vis. Pattern Recog.}, 2020, pp. 2157--2167.

\bibitem{zhang2018residual}
Y.~Zhang, Y.~Tian, Y.~Kong, B.~Zhong, and Y.~Fu, ``Residual dense network for
  image super-resolution,'' in \emph{IEEE Conf. Comput. Vis. Pattern Recog.},
  2018, pp. 2472--2481.

\bibitem{zou2021learning}
Y.~Zou, Y.~Zheng, T.~Takatani, and Y.~Fu, ``Learning to reconstruct high speed
  and high dynamic range videos from events,'' in \emph{IEEE Conf. Comput. Vis.
  Pattern Recog.}, 2021, pp. 2024--2033.

\bibitem{rehder2016extending}
J.~Rehder, J.~Nikolic, T.~Schneider, T.~Hinzmann, and R.~Siegwart, ``Extending
  kalibr: Calibrating the extrinsics of multiple imus and of individual axes,''
  in \emph{ICRA}.\hskip 1em plus 0.5em minus 0.4em\relax IEEE, 2016, pp.
  4304--4311.

\bibitem{rebecq2019high}
H.~Rebecq, R.~Ranftl, V.~Koltun, and D.~Scaramuzza, ``High speed and high
  dynamic range video with an event camera,'' \emph{IEEE Trans. Pattern Anal.
  Mach. Intell.}, vol.~43, no.~6, pp. 1964--1980, 2019.

\bibitem{liu2020single}
Y.-L. Liu, W.-S. Lai, Y.-S. Chen, Y.-L. Kao, M.-H. Yang, Y.-Y. Chuang, and
  J.-B. Huang, ``Single-image hdr reconstruction by learning to reverse the
  camera pipeline,'' in \emph{IEEE Conf. Comput. Vis. Pattern Recog.}, 2020,
  pp. 1651--1660.

\bibitem{rebecq2018esim}
H.~Rebecq, D.~Gehrig, and D.~Scaramuzza, ``Esim: an open event camera
  simulator,'' in \emph{Conference on Robot Learning}.\hskip 1em plus 0.5em
  minus 0.4em\relax PMLR, 2018, pp. 969--982.

\bibitem{wang2020multi}
X.~Wang, Z.~Sun, Q.~Zhang, Y.~Fang, L.~Ma, S.~Wang, and S.~Kwong,
  ``Multi-exposure decomposition-fusion model for high dynamic range image
  saliency detection,'' \emph{IEEE Trans. Circuit Syst. Video Technol.},
  vol.~30, no.~12, pp. 4409--4420, 2020.

\bibitem{mantiuk2011hdr}
R.~Mantiuk, K.~J. Kim, A.~G. Rempel, and W.~Heidrich, ``Hdr-vdp-2: A calibrated
  visual metric for visibility and quality predictions in all luminance
  conditions,'' \emph{ACM Transactions on Graphics.}, vol.~30, no.~4, pp.
  1--14, 2011.

\bibitem{cui2015detail}
G.~Cui, H.~Feng, Z.~Xu, Q.~Li, and Y.~Chen, ``Detail preserved fusion of
  visible and infrared images using regional saliency extraction and
  multi-scale image decomposition,'' \emph{Optics Communications}, vol. 341,
  pp. 199--209, 2015.

\bibitem{eskicioglu1995image}
A.~M. Eskicioglu and P.~S. Fisher, ``Image quality measures and their
  performance,'' \emph{IEEE Trans. on Communications}, vol.~43, no.~12, pp.
  2959--2965, 1995.

\bibitem{mittal2012making}
A.~Mittal, R.~Soundararajan, and A.~C. Bovik, ``Making a “completely blind”
  image quality analyzer,'' \emph{IEEE Sign. Process. Letters}, vol.~20, no.~3,
  pp. 209--212, 2012.

\bibitem{wang2022exploring}
J.~Wang, K.~C. Chan, and C.~C. Loy, ``Exploring clip for assessing the look and
  feel of images,'' in \emph{Proceedings of the AAAI conference on artificial
  intelligence}, 2023.

\bibitem{ke2021musiq}
J.~Ke, Q.~Wang, Y.~Wang, P.~Milanfar, and F.~Yang, ``Musiq: Multi-scale image
  quality transformer,'' in \emph{Int. Conf. Comput. Vis.}, 2021, pp.
  5148--5157.

\bibitem{yang2022maniqa}
S.~Yang, T.~Wu, S.~Shi, S.~Lao, Y.~Gong, M.~Cao, J.~Wang, and Y.~Yang,
  ``Maniqa: Multi-dimension attention network for no-reference image quality
  assessment,'' in \emph{IEEE Conf. Comput. Vis. Pattern Recog.}, 2022, pp.
  1191--1200.

\bibitem{zhang2023blind}
W.~Zhang, G.~Zhai, Y.~Wei, X.~Yang, and K.~Ma, ``Blind image quality assessment
  via vision-language correspondence: A multitask learning perspective,'' in
  \emph{IEEE Conf. Comput. Vis. Pattern Recog.}, 2023, pp. 14\,071--14\,081.

\bibitem{hosu2020koniq}
V.~Hosu, H.~Lin, T.~Sziranyi, and D.~Saupe, ``Koniq-10k: An ecologically valid
  database for deep learning of blind image quality assessment,'' \emph{IEEE
  Trans. Image Process.}, vol.~29, pp. 4041--4056, 2020.

\bibitem{ying2020patches}
Z.~Ying, H.~Niu, P.~Gupta, D.~Mahajan, D.~Ghadiyaram, and A.~Bovik, ``From
  patches to pictures (paq-2-piq): Mapping the perceptual space of picture
  quality,'' in \emph{IEEE Conf. Comput. Vis. Pattern Recog.}, 2020, pp.
  3575--3585.

\bibitem{kingma2014adam}
D.~P. Kingma and J.~Ba, ``Adam: A method for stochastic optimization,''
  \emph{arXiv preprint arXiv:1412.6980}, 2014.

\bibitem{reinhard2002photographic}
E.~Reinhard, M.~Stark, P.~Shirley, and J.~Ferwerda, ``Photographic tone
  reproduction for digital images,'' in \emph{Proceedings of the 29th Annual
  Conference on Computer Graphics and Interactive Techniques}, 2002, pp.
  267--276.

\bibitem{jocher2023yolo}
G.~Jocher, A.~Chaurasia, and J.~Qiu, ``Yolo by ultralytics,'' \emph{URL:
  https://github. com/ultralytics/ultralytics}, 2023.

\end{thebibliography}

\vfill


\end{document}